\documentclass{article} 
\PassOptionsToPackage{table,xcdraw,dvipsnames}{xcolor} 
\usepackage{iclr2027_conference,times}

\usepackage[utf8]{inputenc} 
\usepackage[T1]{fontenc}    
\usepackage{hyperref}       
\usepackage{url}            
\usepackage{booktabs}       
\usepackage{amsfonts}       
\usepackage{nicefrac}       
\usepackage{microtype}      

\usepackage{graphicx}	
\usepackage{amsmath}	
\usepackage{amssymb}	
\usepackage{booktabs}
\usepackage{mathtools}
\usepackage{times}
\usepackage{microtype}
\usepackage{epsfig}
\usepackage[table,xcdraw,dvipsnames]{xcolor}
\usepackage{caption}
\usepackage{pifont}
\usepackage{bbding}

\usepackage{float}
\usepackage{placeins}
\usepackage{color, colortbl}
\usepackage{stfloats}
\usepackage{enumitem}
\usepackage{tabularx}
\usepackage{xstring}
\usepackage{multirow}
\usepackage{xspace}
\usepackage{url}
\usepackage{subcaption}
\usepackage[hang,flushmargin]{footmisc}
\usepackage[flushleft]{threeparttable} 
\usepackage{pifont}

\usepackage{amsthm,amsmath,amssymb}

\usepackage[mathscr]{euscript}

\usepackage{url}

\usepackage{algorithm}
\usepackage{algorithmic}

\usepackage[utf8]{inputenc} 
\usepackage[T1]{fontenc}    
\usepackage{url}            
\usepackage{booktabs}       
\usepackage{amsfonts}       
\usepackage{nicefrac}       
\usepackage{microtype}      
\usepackage{adjustbox} 
\usepackage{comment}
\usepackage{array}

\definecolor{lightblue}{RGB}{230,242,248}
\definecolor{mydarkblue}{rgb}{0, 0.2, 0.5}
\hypersetup{
    colorlinks=true, %
     linkcolor=black,   
    citecolor=mydarkblue,
}

\usepackage{graphicx} 
\usepackage{wrapfig,lipsum,booktabs}
\usepackage{float} %
\usepackage{MnSymbol}
\usepackage{enumitem}
\usepackage{amsthm,amsmath,amssymb}
\usepackage{wrapfig,lipsum,booktabs}
\usepackage{multirow}
\usepackage{algorithm, algorithmic}
\usepackage{bm}
\usepackage{wasysym}
\usepackage{pifont}

\usepackage{makecell} 

\usepackage{fontawesome} 

\usepackage{soul}

\definecolor{basepurple}{RGB}{217,217,255}
\definecolor{gaincolor}{RGB}{0,115,60}
\definecolor{losscolor}{RGB}{180,35,45}

\definecolor{tablebaseblue}{RGB}{74,137,199}
\definecolor{tablemethodpurple}{RGB}{217,217,255}

\newcommand{\scoregain}[2]{%
  \shortstack{#1\\[-1pt]{\scriptsize\textcolor{gaincolor}{(+#2)}}}%
}
\newcommand{\scoreloss}[2]{%
  \shortstack{#1\\[-1pt]{\scriptsize\textcolor{losscolor}{(-#2)}}}%
}
\renewcommand{\scoregain}[2]{%
  \mbox{#1\,{\scriptsize\textcolor{gaincolor}{(+#2)}}}%
}
\renewcommand{\scoreloss}[2]{%
  \mbox{#1\,{\scriptsize\textcolor{losscolor}{(-#2)}}}%
}

\usepackage{enumitem}

\newlist{paramlist}{description}{1}
\setlist[paramlist]{%
  labelwidth=3.4em,   
  labelsep=0.6em,     
  leftmargin=!,       
  itemsep=1pt,        
  topsep=2pt,
  font=\normalfont,   
}

\newcommand{\R}[1]{{%
    \textbf{%
        \ifstrequal{#1}{1}{\textcolor{red}{R#1}}{%
        \ifstrequal{#1}{2}{\textcolor{blue}{R#1}}{%
        \ifstrequal{#1}{3}{\textcolor{magenta}{R#1}}{%
        \ifstrequal{#1}{4}{\textcolor{teal}{R#1}}{%
                           \textcolor{cyan}{R#1}%
        }}}}%
    }%
}}

\newcommand{\xv}[0]{\boldsymbol{x}}

\theoremstyle{plain}
\newtheorem{theorem}{Theorem}[section]
\newtheorem{proposition}[theorem]{Proposition}

\theoremstyle{definition}

\theoremstyle{remark}

\newcommand{\methodname}{DPDP-ft\xspace}
\newcommand{\methodx}{DPDP-inf\xspace}
\newcommand{\methodfull}{Diffusion Policy-steered Diffusion Planner\xspace}

\title{
DPDP: Diffusion Policy–Steered Diffusion Planner for Offline Reinforcement Learning
}

\iclrfinalcopy

\author{Ruyi An \qquad
Mingyuan Zhou\\
The University of Texas at Austin\\
\texttt{ruyian@utexas.edu, mingyuan.zhou@mccombs.utexas.edu}
}

\begin{document}

\maketitle

\begin{abstract}
In offline reinforcement learning, diffusion models fill two roles with complementary strengths. 
Typically guided by a critic, a diffusion policy is value-aware and favors high-return, behavior-supported actions, but it is short-sighted: it never lays out the states it intends to visit. 
A diffusion planner is far-sighted, synthesizing such a long-horizon trajectory, but it is only behavior-driven at execution, realizing each planned transition using only offline behavior.
Each is strong exactly where the other is weak. We introduce DPDP to bring them together at the planner's execution step, where a planned transition becomes an action.
Reading the planner's inverse-dynamics executor as a posterior over actions: the environment's transition likelihood relates an action to the planned next state, while the behavioral prior determines its weight under the offline data. 
Replacing this prior with a value-optimized diffusion policy preserves the planned transition likelihood and favors valuable actions that can serve the plan. We target the resulting clean-action distribution in two ways. A training-free sampler composes policy and inverse-dynamics scores at matched action noise levels and applies Langevin correction; this route uses a shared action noise schedule and an approximately common terminal prior. A preference objective instead amortizes the transfer into a single diffusion executor. 
Across D4RL, DPDP improves the planner it builds on, staying competitive with reactive diffusion policies on locomotion and leading on long-horizon navigation.
\end{abstract}

\section{Introduction}

At every step, an agent faces two coupled questions: where should it go, and what should it do now to get there? The first calls for a plausible view of the future; the second requires choosing an action whose consequences justify taking it. This distinction loosely echoes deliberation and immediate choice in \cite{kahneman2011thinking}'s account of cognition, and a competent decision needs both at once.
In offline reinforcement learning (RL) \citep{levine2020offline}, diffusion models answer these questions separately, as policies or as planners: each answers one well, and neither answers both.

A diffusion policy is the immediate system. It models an expressive, state-conditioned action distribution $\pi_\theta(a_t\mid s_t)$, and a learned critic steers it toward behavior-supported actions with high estimated downstream return \citep{dql, hansen2023idql}: it answers \emph{what to do now}. Yet an action value says nothing about the intermediate states a distant goal must pass through. 
A diffusion planner is the deliberative system. It generates a future state or state--action trajectory \citep{planner,dd}, avoiding the compounding error of one-step rollouts and exposing an explicit path to the goal: it answers \emph{where to go}. To favor promising plans, it can condition generation on a target return, drawing trajectories resembling those that achieved that return in the data. Yet its denoising objective does not perform Bellman backups, so it does not explicitly rank actions that could realize the same planned next state by downstream value.
Each family is thus strong exactly where the other is weak, holding one half of a competent decision. 
Merging the halves is not a matter of mixing outputs. The two models are defined over different variables, namely actions at the current state versus joint distributions over future trajectories, so the critic's action-level value has no natural port into the planner. A bridge must carry that value into the place the two decisions meet, the step where a plan becomes an action, rather than into a policy standing beside the planner.

This step offers the port we need to close the gap. To execute its plan, a planner turns the first planned transition into an action with an inverse-dynamics model, which is itself a distribution over the current action: a posterior. Its likelihood is the planned transition, the deliberative target for where to go next and its prior is the offline behavior policy, the default action an agent would take before any value is weighed. Read this way, the immediate system belongs in the prior. A value-optimized diffusion policy is exactly that behavior prior improved by the critic, and the log-ratio between the two is a preference over actions, a measure of how much value optimization favors each. Transferring value is then a single Bayesian move: hold the likelihood fixed and replace the behavioral prior with the value policy, updating the posterior so the executed action keeps the planner's intended transition while tilting toward what the critic prefers. The action the agent takes becomes the posterior that fuses both, the deliberated plan entering through the likelihood and the immediate value through the prior, on one decision rather than through a scalar proxy. Because the executor and the value policy are diffusion models over the same action space, the transfer rests on two conditions we arrange by construction: a shared noise schedule and a shared terminal prior.

We name the resulting approach DPDP (\methodfull): an executor in which a diffusion policy and a diffusion planner act as one. 
We realize it in two ways that target the same execution distribution:
The first, \methodx, is a training-free, inference-time algorithm composing the two models' action scores at matched noise levels and correcting the samples during sampling, with both models frozen. 
The second, \methodname, finetunes existing inverse dynamics once, amortizing the transfer into a single executor with a preference objective, so deployment is a single reverse pass with no critic in the loop.
On D4RL, both realizations improve the base planner in average return across locomotion, AntMaze, and Maze2D. They narrow its gap to the reactive policy on locomotion and exceed both base models in average return on AntMaze and Maze2D. These results show that value-informed action selection can improve the execution of a state plan without discarding its proposed transition.

Our contributions can be summarized as:
\begin{itemize}
    \item We cast a planner's execution step as a Bayesian action posterior and recast value transfer as a prior replacement: with the planned transition held as the likelihood, we swap the behavioral prior for a value-optimized diffusion policy. Value-informed action selection thus enters the planner's own execution step. 
    To the best of our knowledge, this work is the first method in which a diffusion policy and a diffusion planner jointly shape the executed action.
    \item 
    We develop two realizations of the resulting target: \methodx composes frozen diffusion scores with inference-time correction, while \methodname amortizes the policy-induced preference through executor fine-tuning.
    \item We show that both realizations improve a diffusion planner across the D4RL suites studied and, on AntMaze and Maze2D, outperform both the planner and reactive policy bases in suite-average return.
\end{itemize}

\section{Background and Related Work}
\label{sec:background}

\noindent \textbf{Offline decision-making.} 
Consider a discounted Markov decision process (MDP) defined by a tuple $\mathcal{M}=(\mathcal{S},\mathcal{A},P,r,\rho_0,\gamma)$, where $\mathcal{S}$ is the state space,
$\mathcal{A}\subseteq\mathbb{R}^{d_a}$ is the action space, $P(s' \mid s,a)$ is the transition kernel, $r(s,a)$ is the reward function on each transition, $\rho_0$ is the initial-state distribution, and $\gamma\in[0,1)$ is the discount factor \citep{sutton1998reinforcement,sac}.
A policy $\pi(a\mid s)$ seeks to maximize the expected discounted return $J(\pi)=\mathbb{E}_{\pi,P}[\sum_{t=0}^{\infty}\gamma^t r(s_t,a_t)]$.
Offline reinforcement learning learns such a policy from a fixed dataset $\mathcal{D}=\{\tau^{(n)}\}_{n=1}^{N}$, where each episode $\tau^{(n)}=(s_0,a_0,r_0,s_1,\ldots)$ consists of ordered transition tuples, without further interaction with the environment \citep{fujimoto,levine2020offline,d4rl}.
The learner must therefore represent potentially high-dimensional and multimodal behavior distributions while improving upon the behavior policy without exploiting actions unsupported by the data.
Long-horizon tasks further require reasoning about delayed consequences and trajectory-level structure.
These challenges motivate expressive models capable of representing complex action and trajectory distributions.

\noindent\textbf{Diffusion models.}
Diffusion models have emerged as a powerful class of generative models capable of representing complex distributions through iterative denoising
\citep{diffusion,ddpm,scoredbased2}. Given a clean sample $\xv^0\sim \mathcal P$, a forward process corrupts it over $K$ steps with the Gaussian kernel $q(\xv^k\mid\xv^0)=\mathcal{N}\!(\sqrt{\bar{\alpha}_k}\xv^0,(1-\bar{\alpha}_k)\mathbf{I})$ under a schedule $\{\bar\alpha_k\}_{k=1}^{K}$ with $\bar\alpha_k$ decreasing to $\approx 0$, so that the terminal marginal converges to a data-independent prior $\nu:=\mathcal N(\mathbf 0,\mathbf I)$\footnote{We use superscripts $^{1,\ldots, k}$ to index diffusion steps and subscripts $_{1,\ldots, t}$ to index environment steps.}.
The resulting marginal is $p_k(\xv^k)=\int q(\xv^k\mid\xv^0) p_0(\xv^0)\,\mathrm d\xv^0$, with $p_0=\mathcal P$. If $\bar\alpha_K=0$, its terminal marginal is exactly $\nu=\mathcal N(\mathbf0,\mathbf I)$ for any $\mathcal P$.
A noise-prediction network \(\epsilon_\theta\) is trained by denoising score matching,
\begin{align}
    \mathcal{L}^{\text{(diffusion)}}(\theta) 
    \;=\;
    \mathbb{E}_{\xv^0\sim\mathcal{P},\;k,\;\boldsymbol{\epsilon}\sim\mathcal N(\mathbf 0,\mathbf I)}
    \big[\,\|\boldsymbol{\epsilon} - \epsilon_{\theta}(\xv^k,k)\|_2^{2}\,\big],
    \quad \xv^k=\sqrt{\bar\alpha_k}\,\xv^0+\sqrt{1-\bar\alpha_k}\,\boldsymbol{\epsilon},
    \label{eq:diff_obj}
\end{align}
which recovers the score of the noised marginal through $\nabla_{\xv^k}\log p_k(\xv^k)\approx-\epsilon_\theta(\xv^k,k)/\sqrt{1-\bar\alpha_k}$ \citep{scorebased}. Sampling then starts from $\xv^K\sim\nu$ and iteratively applies $\epsilon_\theta$ to denoise back to a clean sample.
A trained diffusion model is thus determined by two design choices: its noise schedule $\{\bar\alpha_k\}$ and its terminal prior $\nu$. Two models that share both, over the same action representation, define scores on the same noisy action space at matched noise levels, even though their noised marginals $p_{I,k},p_{Q,k},p_{B,k}$ differ; this is what enables us to combine their scores level by level.

Current work applying diffusion models to offline RL mainly assigns two roles \citep{survey}: \textit{i)} serving as \textbf{policies} to support complex multimodal-distribution modeling and \textit{ii)} generating long-term trajectories and executing like \textbf{planners}.

\noindent \textbf{Diffusion Policy} models an expressive, potentially multimodal action distribution $\pi_\theta(a\mid s)$ \citep{dql,chi2025diffusion} in place of the restricted actor distributions commonly used in offline RL \citep{cql,iql,fujimoto,imitation_diffusion}. 
Starting from a diffusion behavior model trained on offline state-action pairs using Eq.~\ref{eq:diff_obj}, a learned critic can favor high-value actions by selecting or guiding samples \citep{sfbc,hansen2023idql,qgdpo}, or by shaping diffusion policy training \citep{dql}. 
Behavior modeling encourages actions supported by the offline data, while the Bellman-trained critic estimates their downstream returns. Thus, even when several actions are plausible under the data, the critic can favor one with higher estimated value.
The policy does not, however, explicitly generate a route through intermediate states toward a distant goal or temporal constraint.

\noindent \textbf{Diffusion Planner} models the joint distribution of future states, or state--action trajectories, and generates a plan from which an action is executed or recovered through inverse dynamics \citep{planner,dd,adaptdiffuser,hierarchical_plan,hd,veteran}. When the current state and a goal are specified, holding them fixed during denoising casts trajectory generation as a structured inpainting problem; return conditioning or reward guidance can further favor desirable plans. Generating a trajectory jointly avoids the compounding rollout errors of one-step dynamics models and supports long-horizon reasoning in sparse-reward and goal-reaching tasks. Nevertheless, a high-return sample describes one possible continuation; its return does not, by itself, estimate the expected downstream value of executing its first action across possible outcomes. This motivates complementing trajectory-level guidance with a critic $Q(s_t,a_t)$ that evaluates candidate actions by their downstream consequences.

These complementary strengths motivate a policy that combines the planner's view of where the agent should go with the critic's assessment of what action it should take now. The diffusion planner supplies plausible, task-directed future trajectories, while the $Q$-guided diffusion policy favors behavior-supported actions with high estimated downstream return. What the literature still lacks is a way to carry the critic's action-level preference into the planner's execution step itself, rather than into a separate policy, and that is the gap we address.

\section{Methodology}
Combining a diffusion planner with a value-optimized diffusion policy hinges on the execution step, the point where a plan becomes an action. The planner proposes a next-state transition, and an inverse-dynamics executor chooses an action to realize it. Trained on offline data, that executor inherits the behavior policy's action preferences; a value-trained policy improves those preferences but says nothing about the proposed transition. We build an executor that respects the planned transition while favoring the actions the value-trained policy prefers.

We build this executor in three steps. First, we cast inverse dynamics
as a conditional diffusion posterior and expose its state-conditioned
null branch as a behavioral prior
(Sec.~\ref{sec:diffusion_for_idm}). Next, we read that posterior
through Bayes' rule and transfer value by replacing its behavioral
prior with the value-optimized policy, an update we show is equivalently
a preference optimization (Sec.~\ref{sec:qpref}). Finally, we realize
the updated posterior in two ways that mirror these two readings: by
composing the models' scores and correcting samples at inference, or by
amortizing the same preference into a single diffusion policy
(Sec.~\ref{sec:realize}).

\subsection{Probabilistic Inverse Dynamics for Plan Execution}
\label{sec:diffusion_for_idm}
A state-trajectory planner proposes
$\hat{\boldsymbol{\tau}}_t =(s_t,\hat s_{t+h},\ldots,\hat s_{t+H})$ where $h$ is the planning stride.
To execute the plan, an inverse dynamics model (IDM) must choose an action for its first proposed transition, from $s_t$ to $g:=\hat s_{t+h}$. State-based diffusion planners commonly use a deterministic inverse model $a_t=f_\phi(s_t,g)$, trained by action regression on offline transitions \citep{dd,adaptdiffuser}.

A single predicted action need not represent the choices available
for a transition. In offline RL, different actions can lead to similar next states
because of redundant actuation, stochastic dynamics, or variation in
the behavior that collected the data \citep{dql}. When these choices form
distinct modes, squared-error regression tends toward their
conditional mean, which may itself be an unlikely action. We instead
model the conditional action distribution with a diffusion-based inverse
dynamics model.

To strengthen adherence to the planned transition, we apply classifier-free guidance (CFG)~\citep{cfg} to the diffusion inverse dynamics model.
For each offline transition $(s_t,a_t,s_{t+h})$, we train the denoiser $\epsilon_\phi(a_t^k,k,s_t,c)$ using Eq.~\ref{eq:diff_obj}. The conditioning variable $c$ here is $s_{t+h}$ for a fraction of training examples and $\varnothing$ otherwise by CFG dropout, where the dropout is independent of the transition. We always retain $s_t$, which specifies the current control context. Thus the same denoiser learns two branches:
a transition-conditioned inverse dynamics model and a
state-conditioned null model. We express $p_I$, the inverse dynamics, and $p_B$, the behavior prior, as 
\begin{align}
p_I(a\mid s,g)
=p_{\mathcal D}(a\mid s,s'=g),\quad
p_B(a\mid s)
=p_{\mathcal D}(a\mid s).
\label{eq:idm_branches}
\end{align}
The learned conditional and null branches approximate $p_I$ and $p_B$, respectively. In particular, the null branch represents behavior at the current state rather than the global action marginal; it acts as a behavior-cloning diffusion policy. 
Thus, $p_B$ serves as a behavioral reference for comparison with another policy $\pi(a\mid s)$ trained on the same data.

At inference, the executor draws its action from the transition-conditioned branch $p_I(a\mid s,g)$ to realize the planner's proposed transition, while the null branch supplies the behavioral reference~$p_B(a\mid s)$.

\subsection{Value-Guided Execution of Planned Transitions}
\label{sec:qpref}
The planner proposes a next state $g$, and the executor must choose an action that can reach it from $s$. The transition-conditioned distribution $p_I(a\mid s,g)$ and its null counterpart $p_B(a\mid s)$ are, respectively, the conditional and marginal action distributions induced by the offline transitions. For a target $g$ with nonzero behavior-induced transition density,
Bayes' rule expresses $p_I$ as the posterior over behavior actions
conditioned on the transition from $s$ to $g$:
\begin{align}
p_I(a\mid s,g)
&=\frac{P(g\mid s,a)\,p_B(a\mid s)}{P(g\mid s)}, \qquad
P(g\mid s)=\int P(g\mid s,a')p_B(a'\mid s)\,da'.
\label{eq:idm_bayes}
\end{align}
Here, $P(g\mid s,a)$ is the environment's transition likelihood measuring how likely action $a$ is to produce the planned next state. Although we cannot evaluate it directly, Eq.~\ref{eq:idm_bayes} identifies its action-dependent~part:
\begin{align}
P(g\mid s,a)
=P(g\mid s)\frac{p_I(a\mid s,g)}{p_B(a\mid s)}
\propto_a \frac{p_I(a\mid s,g)}{p_B(a\mid s)}
\approx \frac{p_\phi(a\mid s,g)} {p_\phi(a\mid s,\varnothing)}.
\label{eq:idm_likelihood_ratio}
\end{align}
The conditional and null branches of the same diffusion-based inverse dynamics model represent the two distributions in this ratio (Sec.~\ref{sec:diffusion_for_idm}).

Learning by behavioral cloning only the offline data, the planner executes its proposed transition by sampling from
\begin{align}
p_I(a\mid s,g)\propto_a P(g\mid s,a)\,p_B(a\mid s).
\label{eq:planner_executor}
\end{align}
The likelihood favors actions that reach $g$, while the behavior prior weights them according to their frequency in the offline data. However, if several actions can plausibly realize the planned transition, their relative preference is not explicitly determined by critic-estimated value. 
On the other hand, a critic-trained diffusion policy $p_Q(a\mid s)$ provides such a preference: its training favors actions with higher learned $Q$-values while retaining a behavior-cloning objective~\citep{dql}.

Because $p_Q$ and $p_B$ are both fit to the same offline data under the same behavior-cloning objective, with Diffusion-QL adding only a value term on top of it, they empirically have approximately the same action support. This makes the two priors directly comparable, and
we can replace the behavior prior in Eq.~\ref{eq:planner_executor} with $p_Q$ while retaining the transition likelihood \citep{ericxing_priorswap}. 
The resulting executor is
\begin{align} p^*(a\mid s,g) &\propto_a P(g\mid s,a)\,p_Q(a\mid s) = p_I(a\mid s,g) \frac{p_Q(a\mid s)}{p_B(a\mid s)}. \label{eq:posterior_update} 
\end{align} 
It remains conditioned on the planned next state through the transition likelihood, while the critic-trained policy replaces behavior frequencies as the action prior. Thus, $p^*$ depends on $g$ even though $p_Q$ does not.

In terms of the learned models, its clean-action target is represented by
\begin{align}
p^*_{\phi}(a\mid s,g)
\propto_a
p_\phi(a\mid s,g)\,
\frac{p_Q(a\mid s)}
     {p_\phi(a\mid s,\varnothing)}.
\label{eq:modeled_posterior}
\end{align}
The conditional and null IDM branches represent $p_I$ and $p_B$. This identity holds at the clean-action level. It does not imply that combining the denoisers' scores at intermediate noise levels yields the score of the forward-noised target. We elaborate this further in Sec.~\ref{sec:realize}.

Eq.~\ref{eq:posterior_update} also admits a preference interpretation. At a fixed state, the log-density ratio
\begin{align}
r^*(s,a)
=\log\frac{p_Q(a\mid s)}{p_B(a\mid s)}+C(s)
\label{eq:optimal_r}
\end{align}
measures how much more strongly $p_Q$ favors $a$ than $p_B$ does with the action-independent term $C(s)$ canceling in comparisons. It describes a preference induced by the shift between action distributions, rather than the critic value $Q(s,a)$ itself.

Although the diffusion models do not provide tractable densities for evaluating this ratio directly, its pairwise preference can be elicited from samples. At a fixed state, draw $a^+\sim p_Q(\cdot\mid s)$ and $a^-\sim p_B(\cdot\mid s)$. For two observed actions $a$ and $a'$, the odds of assigning $a$ to $p_Q$ and $a'$ to $p_B$, rather than the reverse, are
\begin{align}
\frac{p_Q(a\mid s)\,p_B(a'\mid s)}
     {p_Q(a'\mid s)\,p_B(a\mid s)}
=\exp\!\big[r^*(s,a)-r^*(s,a')\big].
\label{eq:preference_odds}
\end{align}
Normalizing these odds gives the Bradley--Terry preference model
$P_r(a\succ a'\mid s)=\sigma(r^*(s,a)-r^*(s,a'))$
\citep{bradley1952rank,dpo}. Here, the ``winner'' label identifies the action drawn from $p_Q$. Kindly note that it does not imply that every sampled $a^+$ has a larger $Q(s,a)$ than its paired~$a^-$. Appendix~\ref{app:preference} gives the derivation.

The same preference characterizes the tempered executor. For each $(s,g)$, we favor actions with high~$r^*$ while anchoring the execution distribution to the planner's posterior:
\begin{align}
\max_{\rho(\cdot\mid s,g)}\quad
\mathbb E_{a\sim\rho(\cdot\mid s,g)}[r^*(s,a)]
-\beta D_{\mathrm{KL}}\!\left(
\rho(\cdot\mid s,g)\,\|\,p_I(\cdot\mid s,g)
\right).
\label{eq:preference_objective}
\end{align}
For $\beta>0$ and a finite normalizer, its unique optimum is
\begin{align}
p^*_\beta(a\mid s,g)
&\propto_a p_I(a\mid s,g)
\exp\!\left(\frac{r^*(s,a)}{\beta}\right)
\propto_a p_I(a\mid s,g)
\left(\frac{p_Q(a\mid s)}{p_B(a\mid s)}\right)^{1/\beta},
\label{eq:oracle}
\end{align}
At $\beta=1$, this preference optimum is exactly the full prior replacement in Eq.~\ref{eq:posterior_update}. More generally, the second term has the same Bayesian form with a composed action prior
$p_\beta(a\mid s)\propto_a
p_B(a\mid s)^{1-1/\beta}p_Q(a\mid s)^{1/\beta}$.
For $\beta\geq1$, this prior interpolates between $p_Q$ at $\beta=1$ and $p_B$ as $\beta\to\infty$. Thus, the preference objective extends the full replacement by controlling how strongly the value policy changes the executor's prior. Appendix~\ref{app:preference2} proves the optimum.

\subsection{Realizing the Unified Execution Plan with Diffusion Models}
\label{sec:realize}

Sec.~\ref{sec:qpref} defines the full prior replacement in
Eq.~\ref{eq:posterior_update} and its extension in
Eq.~\ref{eq:oracle}. The clean-action composition
in Eq.~\ref{eq:modeled_posterior} motivates combining pretrained
diffusion models at inference. The KL-regularized preference
formulation in Eq.~\ref{eq:preference_objective} motivates amortizing
the update into a single executor reverse chain. We develop both
realizations below.

\noindent\textbf{Composed score path.}
The inverse dynamics model and Diffusion-QL policy do not provide
tractable clean-action densities.  Since forward noising does not commute with a product of densities, combining their scores at level $k$ generally does not give the score of $q_k*p^*_\beta$ \citep{rrr_duyilun}. We instead compose their noised marginals. Using a shared action representation and forward kernel $q_k$, let $p_{j,k}=q_k*p_j$ for $j\in\{I,Q,B\}$ and define
\begin{align}
\tilde p_k(a^k\mid s_t,g)
\propto
p_{I,k}(a^k\mid s_t,g)\,
\left(\frac{p_{Q,k}(a^k\mid s_t)}{p_{B,k}(a^k\mid s_t)}\right)^{1/\beta}.
\label{eq:surrogate}
\end{align}
Let $\epsilon_\psi$ denote the denoiser of the Diffusion-QL policy $p_Q$. For exact component scores, this product has the combined noise prediction
\begin{align}
\hat\epsilon^{*}(a_t^k,k)
&=
\epsilon_\phi(a_t^k,k,s_t,g)
+\frac{1}{\beta}\Big[
\epsilon_\psi(a_t^k,k,s_t)
-\epsilon_\phi(a_t^k,k,s_t,\varnothing)
\Big],
\label{eq:composite}
\end{align}
which adds to the transition-conditioned inverse-dynamics prediction the
$Q$-versus-behavior score difference, other arguments suppressed.
At the clean endpoint, Eq.~\ref{eq:idm_bayes} gives
\begin{align}
\tilde p_0(a\mid s,g)
=p^*_\beta(a\mid s,g)
\propto_a
P(g\mid s,a)\,
p_Q(a\mid s)^{1/\beta}
p_B(a\mid s)^{1-1/\beta}.
\label{eq:realized_target}
\end{align}

\begin{proposition}[Composed annealing path]
\label{prop:endpoint}
Suppose the component denoisers give the exact scores of $p_{I,k}$, $p_{Q,k}$, and $p_{B,k}$; the products in Eq.~\ref{eq:surrogate} are normalizable; and $p_{I,K}=p_{Q,K}=p_{B,K}=\nu$. Then Eq.~\ref{eq:composite} represents $\nabla\log\tilde p_k$, with $\tilde p_K=\nu$ and $\tilde p_0=p^*_\beta$. In general, however, $\tilde p_k\ne q_k*\tilde p_0$ at intermediate levels. Thus, the combined field does not generally provide the forward-marginal scores required by an ordinary reverse diffusion chain.
\end{proposition}

Appendix~\ref{app:oracle} proves the proposition and gives a Gaussian case in which the intermediate marginals do agree. Sharing $q_k$ does not establish that $\epsilon_\psi$ scores $q_k*p_Q$. In practice, we use it as an estimate, and $\bar\alpha_K\approx0$ makes the common terminal prior approximate. The clean-endpoint identity remains~algebraic.

\noindent\textbf{Training-free sampling.}
We initialize $a_t^K\sim\nu$ and use the combined prediction
$\hat\epsilon^*$ for each reverse step. Because the intermediate
$\tilde p_k$ are generally not the forward marginals of
$\tilde p_0$, an ordinary reverse chain need not preserve the
composed path. After stepping from $k$ to $j=k-1$, we therefore
apply $M$ Langevin corrections at each reached level $j>0$:
\begin{align}
a_t^j
\leftarrow
a_t^j
-\eta_j
\frac{\hat\epsilon^{*}(a_t^j,j)}
{\sqrt{1-\bar\alpha_j}}
+\sqrt{2\eta_j}\,z,
\qquad z\sim\mathcal N(\mathbf0,\mathbf I).
\label{eq:corrector}
\end{align}
The drift is $\eta_j\nabla\log\tilde p_j$, so each correction
moves the sample toward the composed marginal before the next
reverse step. An ideal annealed sampler with exact, fully mixed
transitions at every level, including the clean endpoint, would
sample $\tilde p_0$ \citep{rrr_duyilun,multicfg}. Our finite-step
sampler corrects only at positive noise levels and finishes with
a denoising step, so its output approximates that target;
unadjusted Langevin also has step-size bias at fixed $\eta_j$.
Each correction evaluates both IDM branches and the Diffusion-QL
denoiser.

\noindent\textbf{Amortized preference alignment.}
The second realization trains a conditional diffusion policy
$\pi_\theta(a\mid s,g)$ toward the same clean target, so deployment
requires only its reverse chain. We initialize $\pi_\theta$ from
the transition-conditioned IDM and keep that IDM frozen as the
reference $p_I$.

Sec.~\ref{sec:qpref} identifies the preference
$r^*(s,a)=\log[p_Q(a\mid s)/p_B(a\mid s)]+C(s)$.
With $p_I$ as the reference, the policy's implicit preference is
\begin{align}
R_{\theta}(a;s,g)
=
\beta\log\frac{\pi_\theta(a\mid s,g)}{p_I(a\mid s,g)}.
\label{eq:outer_reward}
\end{align}
Setting $R_{\theta}(a;s,g)=r^*(s,a)+C(s,g)$ and solving for
$\pi_\theta$,
\begin{align}
\pi_\theta(a\mid s,g)
&\propto_a
p_I(a\mid s,g)
\left(\frac{p_Q(a\mid s)}{p_B(a\mid s)}\right)^{1/\beta}
=p^*_\beta(a\mid s,g)
\quad\text{after normalization}.
\label{eq:amortized_target}
\end{align}
The $Q$-versus-behavior preference thus supplies the value that shifts
the executor from $p_I$ toward $p^*_\beta$.

We obtain preference pairs at a fixed state by drawing
$a^+\sim p_Q(\cdot\mid s_t)$ and
$a^-\sim p_B(\cdot\mid s_t)$, as in
Sec.~\ref{sec:qpref}. These are distributional labels, with no
sample-wise guarantee that $Q(s_t,a^+)>Q(s_t,a^-)$.
For training, the pair shares $(s_t,g)$ and a noise level
$k\sim\mathcal U\{1,\ldots,K\}$, while its actions are corrupted
independently:
$a_\pm^k=\sqrt{\bar\alpha_k}\,a^\pm+
\sqrt{1-\bar\alpha_k}\,\epsilon^\pm$, with
$\epsilon^+,\epsilon^-\overset{\mathrm{iid}}{\sim}
\mathcal N(\mathbf0,\mathbf I)$.

Clean-action likelihood ratios in Eq.~\ref{eq:outer_reward} are
intractable for these diffusion models. Following diffusion
preference optimization \citep{diffusion_dpo,li2026mclr}, we use
differences of denoising errors as per-level surrogates. For
either member $(a,\epsilon)$ of the pair, let
$a^k=\sqrt{\bar\alpha_k}a+
\sqrt{1-\bar\alpha_k}\epsilon$ and define
\begin{align}
D_\theta(a,\epsilon)
=
\|\epsilon-\epsilon_\theta(a^k,k,s_t,g)\|^2
-\|\epsilon-\epsilon_\phi(a^k,k,s_t,g)\|^2.
\label{eq:align_terms}
\end{align}
$D_\theta$ compares the trainable policy with its frozen reference and is
the per-level surrogate for the log-ratio in Eq.~\ref{eq:outer_reward}.
We minimize
\begin{align}
\mathcal L_{\text{align}}(\theta)
=
-\mathbb E\!\left[
\log\sigma\!\left(
-\frac{\beta}{2}
\big[
D_\theta(a^+,\epsilon^+)
-
D_\theta(a^-,\epsilon^-)
\big]\right)\right].
\label{eq:align}
\end{align}
With exact clean-action likelihood ratios and a realizable policy,
the corresponding pairwise objective has
Eq.~\ref{eq:amortized_target} as its population optimum; the
derivation appears in Appendix~\ref{appendix:training_obj}.
Eq.~\ref{eq:align} replaces those ratios with a
single-timestep denoising surrogate, so it targets that optimum
without guaranteeing exact recovery. The uniform timestep
weighting shown here makes $\beta$ an effective preference-strength
parameter; diffusion-ELBO coefficients would give a differently
weighted loss. All reported runs minimize this loss together with a
behavior-cloning anchor to the reference $p_I$, with training objective:
\begin{align}
\mathcal L(\theta)
=\mathcal L_{\text{align}}(\theta)+\lambda\,\mathcal L_{\text{BC}}(\theta),
\qquad
\mathcal L_{\text{BC}}(\theta)
=\mathbb E_{(s_t,a,g),\,k,\,\epsilon}
\big\|\epsilon-\epsilon_\theta(a^k,k,s_t,g)\big\|^2,
\label{eq:align_deployed}
\end{align}
where $\mathcal L_{\text{BC}}$ is the goal-conditioned denoising loss used to 
train the executor and $\lambda$ is the anchor weight. At deployment, only $\pi_\theta$ is used,
without Langevin corrections or calls to the frozen models.

We provide the algorithmic description of both implementations in Appendix~\ref{sec:app:algo}.

\section{Experiments}
\label{sec:exp}

\noindent \textbf{Dataset.}
We evaluate on the D4RL Gym-MuJoCo locomotion, AntMaze, and Maze2D datasets \citep{d4rl}. We report D4RL-normalized returns from closed-loop rollouts of each method's policy, averaged over five random seeds. For every seed we roll out $50$ parallel environments and run, per environment, $3$ episodes on the dense-reward locomotion tasks and $20$ on the sparse-reward AntMaze and Maze2D suites. Each locomotion score therefore aggregates $50{\times}3{\times}5{=}750$ rollouts, and each AntMaze or Maze2D score $50{\times}20{\times}5{=}5000$ rollouts ($1000$ per seed); the larger budget on the binary-success navigation suites with sparse reward resolves per-seed success to $0.1\%$ and makes the reported standard errors meaningful.

\noindent \textbf{Baseline Methods.}
We compare against behavior cloning (BC) \citep{ross2011reduction}, Gaussian policies (CQL \citep{cql}, IQL \citep{iql}), diffusion policies (DQL \citep{dql}, SfBC \citep{sfbc},  IDQL \citep{hansen2023idql}, QGPO \citep{qgdpo}, SRPO \citep{srpo}), and diffusion planners (Diffuser \citep{planner}, AD \citep{adaptdiffuser}, DD \citep{dd}, HD \citep{hd}, DV \citep{veteran}). For these literature baselines, we use their best reported results. We show both DQL's original results and its CleanDiffuser replication \citep{dong2024cleandiffuser}, denoted DQL$^*$. The latter generally performs better in our comparison and serves as the basis for our adapted DQL$+$.

\noindent \textbf{Implementation Details.}
We build \methodname and \methodx on DQL and DV because their components supply the two signals our method combines. DQL learns a diffusion action policy shaped by a $Q$-function, providing value-informed action preferences; DV generates state plans and uses a diffusion inverse dynamics model to execute their transitions \citep{dql,veteran}. Both are competitive baselines, and DV's diffusion executor gives us a direct way to condition action selection on the planner's proposed next state.

Score composition requires the policy and inverse dynamics model to use the same action diffusion process. We therefore implement DQL and DV's inverse dynamics with a shared $10$-step variance-preserving DDPM schedule \citep{ddpm,edm}. Because this differs from the configurations underlying their published scores, we report our schedule-matched adaptations separately as DQL$+$ and DV$+$. For DV$+$, we train the inverse dynamics model with random next-state conditioning dropout, yielding the conditional and null branches used in Sec.~\ref{sec:diffusion_for_idm}; execution uses only the transition-conditioned branch.
Finally, we rescale the shared schedule to give $\bar\alpha_K\approx0$; the three forward marginals then approach the common terminal prior $\mathcal N(\mathbf0,\mathbf I)$. 

\methodx combines the frozen DQL and inverse-dynamics denoisers at matched noise levels, applying $M$ Langevin corrections after each reverse step to a positive noise level ($M{=}1$ by default). It requires no additional training. \methodname instead fine-tunes the executor toward the same target using preference pairs: preferred actions come from the DQL policy, and dispreferred actions from the IDM's behavioral branch. The transition-conditioned IDM serves as the frozen reference, while the conditional-to-null adjustment retains the chosen level of planning guidance.
We implement both methods in CleanDiffuser \citep{dong2024cleandiffuser} with PyTorch \citep{pytorch} and train on a single NVIDIA RTX 6000 Ada GPU. All diffusion models use Adam \citep{adam} and a batch size of $128$. We train the DV planner and DQL policy for $10^6$ gradient steps each, and the IDM for $2\times10^5$ steps. \methodname fine-tunes the executor for another $2\times10^5$ steps at the same batch size. We report results over five random seeds. Appendix~\ref{sec:hyper} provides the full {hyperparameters}, including the value-transfer strength and corrector budget.

\begin{table*}[t!]
    \centering
    \vspace{-7pt}
    \caption{\textbf{Long-horizon navigation} on D4RL. Normalized return averaged over five seeds. The blue \textcolor{tablebaseblue}{DQL+} (value policy) and \textcolor{tablebaseblue}{DV+} (planner) are our reimplemented schedule-matched base models; the purple \methodx{} and \methodname{} are the two realizations of the transfer, each annotated with its gain over the base planner DV+. On both suites, the policy and the planner are individually comparable, and unifying them exceeds {both} on the suite average.}
        \vspace{-7pt}
    \begin{adjustbox}{center}
    \resizebox{\textwidth}{!}{%
    \begin{tabular}{l*{16}{c}*{2}{>{\columncolor{tablemethodpurple}}c}}
        \toprule
        \multirow{2}{*}{\textbf{Dataset}}
        & \multicolumn{1}{c}{\textbf{}}
        & \multicolumn{2}{c}{\textbf{Gaussian policies}}
        & \multicolumn{7}{c}{\textbf{Diffusion policies}}
        & \multicolumn{8}{c}{\textbf{Diffusion planners}} \\
        \cmidrule(lr){2-2}
        \cmidrule(lr){3-4}
        \cmidrule(lr){5-11}
        \cmidrule(lr){12-19}
        & \textbf{BC}
        & \textbf{CQL}
        & \textbf{IQL}
        & \textbf{SfBC}
        & \textbf{DQL}
        & \textbf{DQL$^\ast$}
        & \textcolor{tablebaseblue}{\textbf{DQL+}}
        & \textbf{IDQL}
        & \textbf{QGPO}
        & \textbf{SRPO}
        & \textbf{Diffuser}
        & \textbf{AD}
        & \textbf{DD}
        & \textbf{HD}
        & \textbf{DV}
        & \textcolor{tablebaseblue}{\textbf{DV+}}
        & \textbf{\methodx}
        & \textbf{\methodname} \\
        \midrule
        Antmaze-M-P
        & 0.0 & 14.9 & 71.2 & 81.3
        & 76.6 & \textbf{87.3} & 86.0 & 84.5 & 83.6 & 80.7
        & 6.7 & 12.0 & 8.0 & -- & \textbf{89.0} & 84.0
        & \scoregain{84.5}{0.5} & \scoregain{84.8}{0.8} \\

        Antmaze-M-D
        & 0.0 & 15.8 & 70.0 & 82.0
        & 78.6 & 82.6 & 87.3 & 84.8 & 83.8 & 75.0
        & 2.0 & 6.0 & 4.0 & \textbf{88.7}
        & 87.4 & 82.9
        & \scoregain{86.0}{3.1} & \scoregain{83.5}{0.6} \\

        Antmaze-L-P
        & 0.0 & 53.7 & 39.6 & 59.3
        & 46.4 & 81.3 & 80.9 & 63.5 & 66.6 & 53.6
        & 17.3 & 5.3 & 0.0 & --
        & 76.4 & 79.5
        & \scoreloss{78.0}{1.5}
        & \scoregain{\textbf{92.8}}{13.3} \\

        Antmaze-L-D
        & 0.0 & 61.2 & 47.5 & 45.5
        & 56.6 & 70.6 & 71.2 & 67.9 & 64.8 & 53.6
        & 27.3 & 8.7 & 0.0 & \textbf{83.6}
        & 80.0 & 78.3
        & \scoregain{81.0}{2.7} & \scoregain{81.7}{3.4} \\

        \midrule
        \textbf{Average}
        & 0.0 & 36.4 & 57.1 & 67.0
        & 64.6 & 80.5 & 81.4 & 75.2 & 74.7 & 65.7
        & 13.3 & 8.0 & 3.0 & --
        & 83.2 & 81.2
        & \scoregain{82.4}{1.2}
        & \scoregain{\textbf{85.7}}{4.5} \\

        \midrule
        Maze2D-U
        & 3.8 & 5.7 & 47.4 & 73.9
        & -- & 140.6 & 137.6 & 57.9 & -- & --
        & 113.9 & 135.1 & -- & 128.4 & 136.6 & 129.2
        & \scoregain{142.3}{13.1}
        & \scoregain{\textbf{145.7}}{16.5} \\

        Maze2D-M
        & 30.3 & 5.0 & 34.9 & 73.8
        & -- & 152.0 & 151.3 & 89.5 & -- & --
        & 121.5 & 129.9 & -- & 135.6 & 150.7 & 146.3
        & \scoregain{\textbf{155.5}}{9.2}
        & \scoregain{152.7}{6.4} \\

        Maze2D-L
        & 5.0 & 12.5 & 58.6 & 74.4
        & -- & 186.8 & 185.8 & 90.1 & -- & --
        & 123.0 & 167.9 & -- & 155.8 & 203.6 & 203.1
        & \scoregain{\textbf{209.0}}{5.9}
        & \scoregain{203.7}{0.6} \\

        \midrule
        \textbf{Average}
        & 13.0 & 7.7 & 47.0 & 74.0
        & -- & 159.8 & 158.2 & 79.2 & -- & --
        & 119.5 & 144.3 & -- & 139.9 & 163.6 & 159.5
        & \scoregain{\textbf{168.9}}{9.4}
        & \scoregain{167.4}{7.8} \\

        \bottomrule
    \end{tabular}%
    }
    \end{adjustbox}
    \vspace{-21pt}
    \label{tab:d4rl}
\end{table*}

\vspace{-3pt}
\subsection{Experimental Results}
\vspace{-3pt}

Tables~\ref{tab:d4rl} and~\ref{tab:d4rl_loco} report D4RL-normalized returns. 
The shaded DQL$+$ and DV$+$ columns denote our schedule-matched value policy and planner. 
Since our DPDP is still planning-based, we mark the signed differences for the proposed variants relative to DV$+$. 
Our reimplementations remain competitive with their reported counterparts and show no systematic performance degradation from schedule matching: some suite averages improve, while others decline modestly (e.g., DV$+$on Maze2D). We therefore evaluate value transfer against DV$+$ under the matched configuration.

On AntMaze and Maze2D (Table~\ref{tab:d4rl}), the two bases perform similarly: DQL$+$ and DV$+$ score $81.4$ and $81.2$ on AntMaze, and $158.2$ and $159.5$ on Maze2D. Both proposed variants exceed both bases in suite-average return. \methodname leads on AntMaze at $85.7$ ($+4.5$ over DV$+$), with a particularly large gain on AntMaze-Large-Play ($79.5\!\to\!92.8$). \methodx leads on Maze2D at $168.9$ ($+9.4$). The gains across both realizations support the benefit of combining planned transitions with critic-informed action selection, although the preferred realization differs by suite. 
Combining the planner’s trajectory guidance with the reactive policy’s value-informed action preferences yields higher average returns than either base model on AntMaze and Maze2D.
Locomotion in Table~\ref{tab:d4rl_loco} are dense-reward control tasks with no end goal to reach. 
Here DQL$+$ averages $88.8$, ahead of DV$+$ at $83.4$. Value transfer improves the planner to $84.5$ with \methodx and $84.3$ with \methodname, while the simple, reactive DQL$+$ retains the lead. \methodx improves DV$+$ on every locomotion task, and \methodname does so on the large majority. We observed that value-informed execution helps even where the reactive policy already performs better than the planner, but does not eliminate that gap.

Across both suites, the two variants improve the base planner on all but three of the thirty-two task entries, with two small regressions and one tie. 
They target the execution by implementing differently, \methodx by inference-time score composition and \methodname by amortized preference learning. 
Their returns track each other closely, so the choice between them is a deployment tradeoff: a single amortized pass with no critic at test time, or a training-free sampler that adds correction steps in place of a training stage. On end goal-oriented environments, \methodx achieved better performance on Maze2D while \methodname achieved better performance on AntMaze.

\subsection{Ablation Studies}

\noindent\textbf{$Q$-reranking.}
A simple way to add value to the executor is to sample
$\{a_1,\ldots,a_N\}\sim p_I(\cdot\mid s,g)$ and execute
$a^\star=\arg\max_i Q(s,a_i)$. This tests whether our transfer improves on test-time critic search over IDM proposals. $Q$-reranking improves locomotion performance but yields only modest gains on Maze2D, where both \methodname and \methodx outperform it on every layout (Appendix~\ref{app:qrank}). Its limitation is that value enters only after the candidates are generated. If a useful action has low probability under $p_I$, finding it may require many costly IDM samples and critic evaluations; if it lies outside the proposal support, reranking cannot find it at all. Searching more candidates can also amplify the notorious overestimation issue in the learned $Q$-function \citep{fujimoto}. In short, reranking spends test-time compute to select from a fixed behavior-cloned proposal, whereas the transfer reshapes the executor's distribution itself.

\begin{table*}[!t]
    \centering
    \caption{\textbf{Gym-MuJoCo locomotion} on D4RL. Normalized return averaged over five seeds; columns as in Table~\ref{tab:d4rl}. The blue \textcolor{tablebaseblue}{DQL+} (value policy) and \textcolor{tablebaseblue}{DV+} (planner) are our reimplemented schedule-matched base models. These are dense-reward, reactive tasks with no goal to reach: every step is scored on the spot, so they reward immediate control rather than foresight. The value policy \textcolor{tablebaseblue}{DQL+} therefore leads and planning is unnecessary, yet the transfer still improves the base planner \textcolor{tablebaseblue}{DV+} on nearly every task, narrowing the gap to the reactive frontier.}
    \vspace{-8pt}
    \begin{adjustbox}{center}
    \resizebox{\textwidth}{!}{%
    \begin{tabular}{l*{16}{c}*{2}{>{\columncolor{tablemethodpurple}}c}}
        \toprule
        \multirow{2}{*}{\textbf{Dataset}}
        & \multicolumn{1}{c}{\textbf{}}
        & \multicolumn{2}{c}{\textbf{Gaussian policies}}
        & \multicolumn{7}{c}{\textbf{Diffusion policies}}
        & \multicolumn{8}{c}{\textbf{Diffusion planners}} \\
        \cmidrule(lr){2-2}
        \cmidrule(lr){3-4}
        \cmidrule(lr){5-11}
        \cmidrule(lr){12-19}
        & \textbf{BC}
        & \textbf{CQL}
        & \textbf{IQL}
        & \textbf{SfBC}
        & \textbf{DQL}
        & \textbf{DQL$^\ast$}
        & \textcolor{tablebaseblue}{\textbf{DQL+}}
        & \textbf{IDQL}
        & \textbf{QGPO}
        & \textbf{SRPO}
        & \textbf{Diffuser}
        & \textbf{AD}
        & \textbf{DD}
        & \textbf{HD}
        & \textbf{DV}
        & \textcolor{tablebaseblue}{\textbf{DV+}}
        & \textbf{\methodx}
        & \textbf{\methodname} \\
        \midrule
        Walker2d-M
        & 6.6 & 79.2 & 78.3 & 77.9
        & \textbf{87.0} & 86.8 & 86.9 & 82.5 & 86.0 & 84.4
        & 79.6 & 84.4 & 82.5 & 84.0 & 82.8 & 83.4
        & \scoregain{84.5}{1.1} & \scoregain{85.0}{1.6} \\

        Walker2d-M-R
        & 11.8 & 26.7 & 73.9 & 65.1
        & 95.5 & 98.2 & \textbf{98.4} & 85.1 & 84.4 & 84.6
        & 70.6 & 84.7 & 75.0 & 84.1 & 85.0 & 85.1
        & \scoregain{86.8}{1.7} & \scoregain{86.1}{1.0} \\

        Walker2d-M-E
        & 6.4 & 111.0 & 109.6 & 109.8
        & 110.1 & 111.6 & 110.8 & 112.7 & 110.7 & \textbf{114.0}
        & 106.9 & 108.2 & 108.8 & 107.1 & 109.2 & 108.6
        & \scoregain{110.0}{1.4} & \scoregain{110.7}{2.1} \\

        Hopper-M
        & 29.0 & 58.0 & 66.3 & 57.1
        & 90.5 & 96.5 & 97.0 & 65.4 & 98.0 & 95.5
        & 74.3 & 96.6 & 79.3 & \textbf{99.3}
        & 83.6 & 83.2
        & \scoregain{84.3}{1.1} & \scoregain{83.2}{0.0} \\

        Hopper-M-R
        & 11.3 & 48.6 & 94.7 & 86.2
        & 101.3 & \textbf{101.6} & 101.4 & 92.1 & 96.9 & 101.2
        & 93.6 & 92.2 & 100.0 & 94.7 & 91.9 & 91.2
        & \scoregain{92.2}{1.0} & \scoregain{91.9}{0.7} \\

        Hopper-M-E
        & 111.9 & 98.7 & 91.5 & 108.6
        & 111.1 & 111.1 & 108.7 & 108.6 & 108.0 & 100.1
        & 103.3 & 111.6 & 111.8 & \textbf{115.3}
        & 110.0 & 109.8
        & \scoregain{110.7}{0.9} & \scoregain{110.4}{0.6} \\

        HalfCheetah-M
        & 36.1 & 44.4 & 47.4 & 45.9
        & 51.1 & 52.3 & 52.4 & 51.0 & 54.1 & \textbf{60.4}
        & 42.8 & 44.2 & 49.1 & 46.7 & 50.4 & 50.8
        & \scoregain{50.9}{0.1} & \scoregain{51.9}{1.1} \\

        HalfCheetah-M-R
        & 38.4 & 46.2 & 44.2 & 37.1
        & 47.8 & 47.9 & 48.0 & 45.9 & 47.6 & \textbf{51.4}
        & 37.7 & 38.3 & 39.3 & 38.1 & 45.8 & 46.0
        & \scoregain{46.8}{0.8} & \scoregain{47.3}{1.3} \\

        HalfCheetah-M-E
        & 35.8 & 62.4 & 86.7 & 92.6
        & \textbf{96.8} & 95.5 & 96.0 & 95.9 & 93.5 & 92.2
        & 88.9 & 89.6 & 90.6 & 92.5 & 92.7 & 92.7
        & \scoregain{94.6}{1.9} & \scoreloss{92.0}{0.7} \\

        \midrule
        \textbf{Average}
        & 31.9 & 63.9 & 77.0 & 75.6
        & 87.9 & \textbf{89.1} & 88.8 & 82.1 & 86.6 & 87.1
        & 77.5 & 83.3 & 81.8 & 84.6 & 83.5 & 83.4
        & \scoregain{84.5}{1.1} & \scoregain{84.3}{0.9} \\

        \bottomrule
    \end{tabular}%
    }
    \end{adjustbox}
        \vspace{-18pt}
    \label{tab:d4rl_loco}
\end{table*}

\noindent\textbf{Value strength.}
We sweep the value strength $\beta$ for both realizations (Appendix~\ref{app:betaw}). On locomotion, where DQL$+$ already leads DV$+$, full value transfer ($\beta{=}1$) adds at most one or two points per task.


\noindent\textbf{Langevin correction.}
We vary \methodx's correction budget and measure inference latency alongside \methodname (Appendix~\ref{app:corrector}). One correction step improves uncorrected composition on every suite. Small budgets provide the strongest gains; larger budgets reduce return on AntMaze while increasing latency. The amortized \methodname remains the faster deployment option.

\vspace{-6pt}
\section{Discussion and Conclusion}
\vspace{-6pt}
\noindent \textbf{Limitations.}
Because diffusion models do not compose exactly \citep{rrr_duyilun, composing_diffusion}, the two realizations also approximate their common clean-action target in different ways. For \methodx, composing scores at each noise level does not generally produce the forward marginals of that target \citep{rrr_duyilun,composing_diffusion}, and finite Langevin correction does not remove all sampling error. Its theoretical characterization further assumes that the $Q$-trained denoiser scores the forward-noised distribution of its own generated actions, which $Q$-guided training does not guarantee. Repeated evaluations of the conditional IDM, null IDM, and DQL denoisers increase inference latency. For \methodname, denoising-error differences substitute for intractable clean-action likelihood ratios, and preference fine-tuning adds a training stage.

\noindent \textbf{Future Work.}
The transfer needs only a value-optimized action prior and a transition-conditioned executor that share a noise schedule, and value enters through the prior alone, which makes it easy to extend. 
A world-model planner whose inverse dynamics conditions on high-dimensional observations would carry value-aware execution to visual control \citep{hafner2020dreamer,universal_policy}, recovering the transition likelihood from the executor's two branches exactly as here. Because value is confined to the prior, a sharper offline critic drops directly into that slot, turning progress on value estimation into better plan execution without changing the method. 
Finally, the value strength $\beta$ is fixed per suite; learning a state-dependent $\beta$ that trusts the critic more where its estimates are reliable \citep{wu2024betadpo} is a natural refinement of the single knob.

\noindent \textbf{Conclusion.}
We presented a way to steer the execution of a diffusion planner by value-informed action preferences. The planner's proposed transition determines which actions are compatible with its plan, while a critic-trained policy changes their relative preference through a tempered prior substitution. We realize this target by composing frozen scores at inference with \methodx or by training a conditional executor through preference alignment with \methodname. On D4RL, both improve the base planner in average return across locomotion, AntMaze, and Maze2D; on the navigation suites, they exceed both base models on average. 
Returning to the question we opened with, the planner still answers \emph{where to go}, and the executor uses value information from the policy to decide \emph{what to do now to get there}: an immediate choice made with deliberate foresight.

\ificlrfinal
\section*{Acknowledgement}
Discussion with Xiang Li of the University of Michigan is highly valued and appreciated.

\bibliography{iclr2027_conference}
\bibliographystyle{iclr2027_conference}

\clearpage

\newpage

\appendix

\renewcommand{\thesection}{\Alph{section}}
\renewcommand\thefigure{A\arabic{figure}}
\renewcommand\thetable{A\arabic{table}}

\setcounter{figure}{0}
\setcounter{table}{0}

\section{Extended Related Work}
\label{app:related}

\subsection{Diffusion Policies}
Diffusion policies replace Gaussian or deterministic actors \citep{cql,iql,fujimoto} with expressive, multimodal action distributions \citep{dql, imitation_diffusion,chi2025diffusion}. To excel beyond pure behavioral cloning, a critic then steers this distribution toward high-value actions, either by selecting or guiding behavior samples \citep{sfbc,hansen2023idql,qgdpo,srpo} or by adding a value term to policy training \citep{dql,kl_reg_rl,gao2025bdpo, zhang2025qipo, fql, rql, ren2025dppo}.
Guidance-based variants \citep{frans2025cfgrl,cfgpolicy} build the improved policy from the score difference of two branches, an optimality-conditioned and an unconditioned policy, which tilts the sampled action away from the behavior policy.
Distillation further reduces inference cost by compressing iterative diffusion sampling into a one-step policy \citep{wang2025onestep,chen2024diffusion}.
In every case, however, the tilt is applied to $\pi(a\mid s)$, conditioned on the current state alone. Where the agent should go next is never represented; long-horizon structure reaches the action only through the scalar $Q$, which becomes less reliable as the horizon grows and rewards become sparse.

\subsection{Diffusion Planners: Progress Concentrates on the Trajectory, Not the Executor}
Diffuser models state-action trajectories and plans by guided inpainting \citep{planner}; Decision Diffuser generates state-only plans and recovers actions with an inverse-dynamics model \citep{dd,idm1,idm2}. 
Recent planners improve how the trajectory is searched: tree search over partially denoised plans \citep{yoon2025mctd}, model-predictive ranking of sampled plans \citep{zhou2025dmpc,balim2026modelbased}, prior-guided sampling \citep{ki2025priorguided}, and stitching of short-horizon segments \citep{luo2025compdiffuser}. 
These methods change which plan is sampled, and value enters them, when it does, by ranking whole plans.
A state plan still has to be turned into control. When the plan specifies states only \citep{dd,veteran}, an inverse-dynamics model trained on dataset transitions $(s,a,s')$ recovers the action for the first planned transition. Because it is fit to behavior data, it implicitly carries the behavior policy as its prior: among the actions that could realize $s\to s'$, it prefers those the dataset took most often, not those with the highest downstream value. Ranking plans does not change this, since the ranking acts before the executor and never compares actions for a fixed transition. The step at which a plan becomes an action therefore has no mechanism for value-based action selection.

\subsection{Combining Planners and Policies}
Diffusion policies and diffusion planners apply the same tool, diffusion models, to the same problem, offline RL, and differ only in what they generate. The two have developed as largely separate lines, and each performs well where the other is weaker: policies where immediate control dominates, planners where long-horizon structure decides success \citep{dd,veteran}. Because both are diffusion models trained on the same data, they share the machinery needed to combine them, which makes the combination natural. Yet, to our knowledge, no prior method combines them so that both shape the action that is executed. We seek this combination to obtain an agent that follows the planner's long-horizon route while choosing, at each step, the action the critic values most among those that realize it.

The connections that do exist are implicit. 
Generated trajectories serve as training data for generative data augmentation \citep{gdadar} where a guided diffusion model synthesizes experience \citep{synthetic_replay,jackson2024policy,lee2024gta,li2024diffstitch}, and a separately trained agent consumes the result. 
The value signal then reaches the executed action only after a second agent is retrained on synthetic data; errors in the generated trajectories become distribution shift for that agent, and every change to the generator requires another round of training. Planners that rank sampled plans with a value estimate \citep{zhou2025dmpc,balim2026modelbased} use value only to choose among trajectories, not among actions.

\section{Further Methodological Discussion}

This appendix proves the claims behind the execution target and its two
realizations, in the order Sec.~\ref{sec:qpref} and
Sec.~\ref{sec:realize} develop them. The starting point is the Bayesian
reading of the executor: $p_I$ is a posterior whose likelihood is the
planned transition and whose prior is the behavior policy, and value
transfer replaces that prior with the value policy. We first show that
the shift between the two priors is a preference
(App.~\ref{app:preference}), then solve the KL-anchored objective and
identify its optimum as the prior-replaced posterior
(App.~\ref{app:preference2}). We assemble these into the amortized
objective and the denoising surrogate we train
(App.~\ref{appendix:training_obj}). Finally, we prove the composed
annealing path that reaches the same target without training
(App.~\ref{app:oracle}).

\subsection{The Preference Induced by the Value Policy}
\label{app:preference}
The transfer in Sec.~\ref{sec:qpref} replaces the behavior prior $p_B$ with the value prior $p_Q$. We show that the shift between the two priors is itself a preference, so the replacement can be reached by preference optimization. Drawing preferred actions from the value prior and dispreferred actions from the behavior prior encodes, in population, the log-density-ratio reward of Eq.~\ref{eq:optimal_r}. Fix a state $s$ and sample $a^+\sim p_Q(\cdot\mid s)$ and $a^-\sim p_B(\cdot\mid s)$ independently. For two distinct actions $a,a'$, realizing the draw as (preferred $a$, dispreferred $a'$) has probability $p_Q(a)p_B(a')$, and the opposite ordering has probability $p_Q(a')p_B(a)$. Conditioning on one of the two orderings occurring, the first is chosen with probability
\begin{align}
\Pr[a\succ a'\mid s]=\frac{p_Q(a)p_B(a')}{p_Q(a)p_B(a')+p_Q(a')p_B(a)}
=\sigma\!\Big(\log\tfrac{p_Q(a)}{p_B(a)}-\log\tfrac{p_Q(a')}{p_B(a')}\Big),
\end{align}
dividing through by $p_Q(a)p_B(a')$, with $\sigma(x)=(1+e^{-x})^{-1}$. A Bradley--Terry model predicts $\sigma(r(s,a)-r(s,a'))$; equating the two on every pair forces $r$ and $\log(p_Q/p_B)$ to differ only by an action-independent term:
\begin{align}
r^*(s,a)=\log\frac{p_Q(a\mid s)}{p_B(a\mid s)}+C(s).
\end{align}
The offset $C(s)$ is unidentifiable from comparisons and cancels wherever the reward enters a difference. The preference the value policy expresses over behavior is thus exactly the log-ratio between the new prior $p_Q$ and the old prior $p_B$, the quantity a preference objective transfers into the executor.

\subsection{The KL-Anchored Optimum as a Prior Replacement}
\label{app:preference2}
We solve Eq.~\ref{eq:preference_objective} and record its guided form. Fix $(s,g)$ and, with the reward $r^*$ of App.~\ref{app:preference}, consider
\begin{align}
\max_{\rho}\ \mathbb E_{a\sim\rho}[r^*(s,a)]-\beta\,D_\mathrm{KL}(\rho\,\|\,p_I),\qquad\beta>0.
\end{align}
Write $\rho_\beta(a)\propto_a p_I(a)\exp(r^*(s,a)/\beta)$ with finite normalizer $Z_\beta$. Collecting terms against $\rho_\beta$ turns the objective into a single divergence,
\begin{align}
\mathbb E_\rho[r^*]-\beta\,D_\mathrm{KL}(\rho\|p_I)=\beta\log Z_\beta-\beta\,D_\mathrm{KL}(\rho\|\rho_\beta),
\end{align}
which is largest, across all $\rho$, only at $\rho=\rho_\beta$. Inserting $r^*=\log(p_Q/p_B)+C(s)$ (the offset merges into the normalizer) yields
\begin{align}
p^*_\beta(a\mid s,g)\propto_a p_I(a\mid s,g)\Big(\frac{p_Q(a\mid s)}{p_B(a\mid s)}\Big)^{1/\beta},
\end{align}
which is Eq.~\ref{eq:oracle}. Read through Bayes' rule on the reference, $p_I\propto_a P(g\mid s,a)p_B$, this optimum leaves the transition likelihood untouched and updates only the prior: the behavior prior is raised toward the value prior, giving the tempered prior $p_Q^{1/\beta}p_B^{1-1/\beta}$ in which $1/\beta$ sets how far the update moves, so $p^*_\beta\propto_a P(g\mid s,a)\,p_Q^{1/\beta}p_B^{1-1/\beta}$. At $\beta=1$ the behavior prior is replaced by $p_Q$ outright, and as $\beta\to\infty$ the update fades and $p^*_\beta\to p_I$.

\subsection{The Amortized Alignment Objective}
\label{appendix:training_obj}
Apps.~\ref{app:preference} and~\ref{app:preference2} give the target: the preference $r^*=\log(p_Q/p_B)$ and the KL-anchored optimum $p^*_\beta$, the posterior whose behavior prior has been replaced by the tempered value prior. Here we recover that optimum from sampled preferences alone, then give the denoising surrogate we actually train. Fix a state--target pair $(s,g)$; every distribution over the action $a$ is conditioned on it, with winners $a^+\sim p_Q(\cdot\mid s)$, losers $a^-\sim p_B(\cdot\mid s)$, and reference $p_I(\cdot\mid s,g)$.

Two results from above drive the amortization. For any reward $r$, the KL-anchored objective of Eq.~\ref{eq:preference_objective} has the Gibbs optimum $\rho_\beta(a)\propto_a p_I(a)\exp(r(s,a)/\beta)$ (App.~\ref{app:preference2}), and the preference the value policy expresses is $r^*(s,a)=\log(p_Q(a\mid s)/p_B(a\mid s))+C(s)$ (App.~\ref{app:preference}), whose offset $C(s)$ cancels in every pairwise difference.

We fit $\pi_\theta$ without estimating $r^*$. Inverting the Gibbs optimum writes the reward implied by any policy as $r(s,a)=\beta\log(\pi_\theta(a)/p_I(a))+\beta\log Z_\beta$, whose constant drops from pairwise differences, so the Bradley--Terry likelihood of the sampled pairs becomes
\begin{align}
\mathcal L_{\mathrm{clean}}(\theta)=-\,\mathbb E\big[\log\sigma\big(R_{\theta}(a^+;s,g)-R_{\theta}(a^-;s,g)\big)\big],
\label{eq:app_clean}
\end{align}
with $R_{\theta}$ of Eq.~\ref{eq:outer_reward}. Setting $R_{\theta}=r^*$, solving for $\pi_\theta$, and using the Bayes factorization $p_I\propto_a P(g\mid s,a)p_B$ (Eq.~\ref{eq:idm_bayes}),
\begin{align}
\pi_\theta(a\mid s,g)\propto_a p_I\Big(\frac{p_Q}{p_B}\Big)^{1/\beta}\propto_a P(g\mid s,a)\,p_Q^{1/\beta}p_B^{1-1/\beta}=p^*_\beta,
\end{align}
exactly Eq.~\ref{eq:amortized_target}. With exact densities the preference objective is minimized precisely at the prior-replaced posterior.

The clean ratio in $R_{\theta}$ is intractable for these diffusion models, so it is replaced by a difference of denoising errors. Under $a^k=\sqrt{\bar\alpha_k}a+\sqrt{1-\bar\alpha_k}\epsilon$, the standard weighted bound gives $-\log\pi_\theta(a)\le\tfrac12\mathbb E_{k,\epsilon}\omega(k)\|\epsilon-\epsilon_\theta(a^k,k,s,g)\|^2+\mathrm{const}$, and identically for $p_I$ with $\epsilon_\phi(\cdot,g)$; subtracting the two bounds and writing $D_\theta$ as in Eq.~\ref{eq:align_terms},
\begin{align}
\log\frac{\pi_\theta}{p_I}\approx-\tfrac12\mathbb E_{k,\epsilon}\omega(k)D_\theta.
\end{align}
Subtracting two variational bounds returns the difference of the bounded log-densities only when their gaps cancel, so these identities are exact at the denoising-score-matching optimum and approximate otherwise. We now make the connection between the trained loss and the exact optimum explicit. Substituting this identity into $R_{\theta}$ (Eq.~\ref{eq:outer_reward}) writes the exact reward margin as a single expectation over noise levels,
\begin{align}
R_{\theta}(a^+)-R_{\theta}(a^-)
=\mathbb E_{k,\epsilon^+,\epsilon^-}[\,m\,],
\qquad
m:=-\tfrac{\beta}{2}\,\omega(k)\big(D_\theta(a^+,\epsilon^+)-D_\theta(a^-,\epsilon^-)\big),
\label{eq:app_margin}
\end{align}
with $D_\theta$ as in Eq.~\ref{eq:align_terms}. The exact objective $\mathcal L_{\mathrm{clean}}$ (Eq.~\ref{eq:app_clean}) applies $\log\sigma$ to this full expectation, whereas the trainable loss Eq.~\ref{eq:align} draws one $(k,\epsilon^+,\epsilon^-)$ per step and applies $\log\sigma$ under the expectation. Since $-\log\sigma$ is convex, Jensen's inequality links the two directly:
\begin{align}
\mathcal L_{\text{align}}(\theta)
=-\,\mathbb E\,\mathbb E_{k,\epsilon}\big[\log\sigma(m)\big]
\ \ge\
-\,\mathbb E\big[\log\sigma(\mathbb E_{k,\epsilon}\,m)\big]
=\mathcal L_{\mathrm{clean}}(\theta).
\label{eq:app_jensen}
\end{align}
The diffusion loss is thus a \emph{variational upper bound} on the exact preference objective, so minimizing Eq.~\ref{eq:align} minimizes $\mathcal L_{\mathrm{clean}}$, whose unique minimizer is $p^*_\beta$. The two minimizers coincide at $p^*_\beta$, when the per-level margin $m$ is constant in $(k,\epsilon)$ and the ELBO gaps of $\pi_\theta$ and $p_I$ cancel, the idealized limit of exact scores and matched weighting; away from it the surrogate targets $p^*_\beta$ up to this Jensen slack. The reported runs minimize this bound together with a behavior-cloning anchor to $p_I$ of weight $\lambda$ (Eq.~\ref{eq:align_deployed}); the derivation here is the $\lambda=0$ case, and $\lambda>0$ biases the trained student toward the reference.

\subsection{Composed Annealing Path}
\label{app:oracle}

We prove Proposition~\ref{prop:endpoint} and clarify why the
composed score cannot generally be used as the score of an ordinary
reverse diffusion process. Fix $(s,g)$ and write
\[
q_k(a^k\mid a^0)
=
\mathcal N\!\left(
a^k;\sqrt{\bar\alpha_k}\,a^0,
(1-\bar\alpha_k)\mathbf I
\right),
\qquad
p_{j,k}=q_k*p_j,\quad j\in\{I,Q,B\}.
\]
Let
\[
w_I=1,\qquad
w_Q=\frac{1}{\beta},\qquad
w_B=-\frac{1}{\beta},
\]
so that $\sum_j w_j=1$. We assume the relevant densities are
positive on their common support and that
$Z_k=\int\prod_j p_{j,k}(a)^{w_j}\,da$ is finite and nonzero at
every level considered. Normalizability is an explicit assumption
because some exponents may be negative.

\paragraph{Score of the composed path.}
Under the exact-score assumption, each frozen denoiser satisfies
\[
\nabla_{a^k}\log p_{j,k}(a^k)
=
-\frac{\epsilon_j(a^k,k)}
{\sqrt{1-\bar\alpha_k}},
\qquad k>0.
\]
Eq.~\ref{eq:composite} is the same weighted sum of the three
noise predictions. Consequently, its score form is
\begin{align}
-\frac{\hat\epsilon^{*}(a^k,k)}
{\sqrt{1-\bar\alpha_k}}
&=
\sum_j w_j\nabla_{a^k}\log p_{j,k}(a^k)
\nonumber\\
&=
\nabla_{a^k}\log
\left[
\frac{1}{Z_k}
\prod_j p_{j,k}(a^k)^{w_j}
\right]
=
\nabla_{a^k}\log\tilde p_k(a^k).
\label{eq:appendix_composed_score}
\end{align}
Thus the composed field is exactly the score of the geometric
marginal $\tilde p_k$, under the proposition's assumptions.

\paragraph{Endpoints.}
If all three terminal marginals equal $\nu$, then
\[
\tilde p_K
\propto
\prod_j\nu^{w_j}
=
\nu^{\sum_jw_j}
=
\nu.
\]
For the Gaussian forward kernel above, $\bar\alpha_K=0$ is a
sufficient condition for this equality for arbitrary clean
distributions. With a conventional finite schedule for which
$\bar\alpha_K$ is small but positive, terminal equality is generally
approximate.

At $k=0$, the forward kernel is the identity, and therefore
\[
\tilde p_0(a\mid s,g)
\propto
p_I(a\mid s,g)\,
p_Q(a\mid s)^{1/\beta}\,
p_B(a\mid s)^{-1/\beta}.
\]
Using the population relation
$p_I(a\mid s,g)\propto_a P(g\mid s,a)p_B(a\mid s)$ gives
\[
\tilde p_0(a\mid s,g)
\propto
P(g\mid s,a)\,
p_Q(a\mid s)^{1/\beta}\,
p_B(a\mid s)^{1-1/\beta},
\]
which is Eq.~\ref{eq:realized_target}. Hence the geometric
marginals connect the common terminal prior to the intended
clean-action target.

\paragraph{When does this path equal forward noising of the target?}
In general, multiplying the noised factors does not give the same
distribution as noising their clean product:
$\tilde p_k\neq q_k*\tilde p_0$. A shared-covariance Gaussian
family is a useful case where equality does hold.

Suppose $p_j=\mathcal N(m_j,\Sigma)$ for all three factors, with
the same positive-definite covariance $\Sigma$. Since
$\sum_jw_j=1$,
\[
\sum_jw_j(a-m_j)^\top\Sigma^{-1}(a-m_j)
=
(a-\bar m)^\top\Sigma^{-1}(a-\bar m)+C,
\qquad
\bar m=\sum_jw_jm_j,
\]
where $C$ does not depend on $a$. After normalization,
$\tilde p_0=\mathcal N(\bar m,\Sigma)$. Forward noising gives
\[
q_k*\tilde p_0
=
\mathcal N\!\left(
\sqrt{\bar\alpha_k}\,\bar m,\,
\Sigma_k
\right),
\qquad
\Sigma_k=
\bar\alpha_k\Sigma+(1-\bar\alpha_k)\mathbf I.
\]
Each noised factor is
$p_{j,k}=\mathcal N(\sqrt{\bar\alpha_k}\,m_j,\Sigma_k)$.
Applying the same weighted-product calculation to these factors
therefore yields
\[
\tilde p_k
=
\mathcal N\!\left(
\sqrt{\bar\alpha_k}\,\bar m,\Sigma_k
\right)
=
q_k*\tilde p_0.
\]

\paragraph{A counterexample.}
Unequal covariances break this equality even when every geometric
marginal is well defined. In one dimension, take
\[
p_I=\mathcal N(0,2),\qquad
p_B=\mathcal N(0,5),\qquad
p_Q=\mathcal N(0,10),\qquad
w=\beta=1.
\]
The weights are $(1,1,-1)$, so $\tilde p_0\propto p_Ip_Q/p_B$.
Its precision is
$\frac12+\frac1{10}-\frac15=\frac25$, giving
$\tilde p_0=\mathcal N(0,\frac52)$. At
$\bar\alpha_k=\frac12$, forward noising of this target has
variance
\[
\operatorname{Var}(q_k*\tilde p_0)
=
\frac12\cdot\frac52+\frac12
=
\frac74.
\]
At the same level, the noised $I$, $Q$, and $B$ factors have
variances $\frac32$, $\frac{11}{2}$, and $3$, respectively.
Their geometric product instead has precision
\[
\frac{1}{3/2}+\frac{1}{11/2}-\frac{1}{3}
=
\frac{17}{33},
\]
and thus variance $\frac{33}{17}\neq\frac74$. The composed field
is the score of $\tilde p_k$, but not the score of
$q_k*\tilde p_0$ at this intermediate level.

\paragraph{Implication for sampling.}
The endpoint result supports annealed sampling: at each level,
a corrector can use Eq.~\ref{eq:appendix_composed_score} to move
samples toward $\tilde p_k$
\citep{rrr_duyilun}. Ideal transitions that
equilibrate at each level, including the clean endpoint, return
samples from $\tilde p_0$. Eq.~\ref{eq:corrector} implements an
unadjusted Langevin approximation. With a fixed step size, even
arbitrarily many updates can retain discretization bias; moreover,
its $\epsilon$-to-score conversion is undefined at $k=0$ as
written. Our finite-step implementation corrects at positive noise
levels and applies a final denoising update. It therefore targets
$\tilde p_0$ approximately, without claiming exact clean-endpoint
sampling at finite cost.

\section{Algorithmic Description of Proposed Algorithm}
\label{sec:app:algo}
The two realizations are summarized in Algorithm~\ref{alg:composition} (\methodx) and Algorithm~\ref{alg:align} (\methodname).

\begin{algorithm}[H]
\caption{\methodx: training-free composed execution}
\label{alg:composition}
\begin{algorithmic}[1]
\STATE {\bfseries Input:} state $s$, planned next state $g$; frozen inverse-dynamics denoiser $\epsilon_\phi$ with null branch $\epsilon_\phi(\cdot,\varnothing)$; Diffusion-QL denoiser $\epsilon_\psi$; schedule $\{\bar\alpha_k\}_{k=1}^{K}$ with step retention $\alpha_k{=}\bar\alpha_k/\bar\alpha_{k-1}$ and reverse variance $\sigma_k^2{=}\tfrac{1-\bar\alpha_{k-1}}{1-\bar\alpha_k}(1-\alpha_k)$; value strength $\beta$; corrector budget $M$; step sizes $\{\eta_j\}$
\STATE Sample $a^{K}\sim\mathcal N(\mathbf 0,\mathbf I)$
\FOR{$k=K$ {\bfseries to} $1$}
\STATE $\hat\epsilon^{*}(a^{k},k)\gets \epsilon_\phi(a^{k},k,s,g)+\tfrac{1}{\beta}\big[\epsilon_\psi(a^{k},k,s)-\epsilon_\phi(a^{k},k,s,\varnothing)\big]$
\STATE Predictor ($k\to j\!=\!k-1$), DDPM reverse step: $a^{j}\gets\dfrac{1}{\sqrt{\alpha_k}}\Big(a^{k}-\dfrac{1-\alpha_k}{\sqrt{1-\bar\alpha_k}}\,\hat\epsilon^{*}(a^{k},k)\Big)+\sigma_k\,z,\ z\sim\mathcal N(\mathbf 0,\mathbf I)$ \ ($z{=}\mathbf 0$ if $j{=}0$)
\IF{$j>0$}
\FOR{$m=1$ {\bfseries to} $M$} 
\STATE $a^{j}\gets a^{j}-\eta_j\,\dfrac{\hat\epsilon^{*}(a^{j},j)}{\sqrt{1-\bar\alpha_j}}+\sqrt{2\eta_j}\,z,\quad z\sim\mathcal N(\mathbf 0,\mathbf I)$ \COMMENT{Langevin correction}
\ENDFOR
\ENDIF
\ENDFOR
\STATE {\bfseries return} $a^{0}$
\end{algorithmic}
\end{algorithm}

\begin{algorithm}[H]
\caption{\methodname: amortized preference alignment}
\label{alg:align}
\begin{algorithmic}[1]
\STATE {\bfseries Input:} inverse-dynamics executor $\phi$ (frozen reference); Diffusion-QL policy $p_Q$; schedule $\{\bar\alpha_k\}$; preference temperature $\beta$; anchor weight $\lambda$
\STATE Initialize student $\theta\gets\phi$
\WHILE{not converged}
\STATE Sample $(s,g)$ from $\mathcal D$
\STATE $a^{+}\sim p_Q(\cdot\mid s)$ \COMMENT{preferred: value policy};\quad $a^{-}\sim p_\phi(\cdot\mid s,\varnothing)$ \COMMENT{dispreferred: behavioral null branch}
\STATE Sample level $k$ and noise $\epsilon$; form $a^{+,k},a^{-,k}$
\STATE $\Delta_\theta(a^{k})\gets\|\epsilon-\epsilon_\theta(a^{k},k)\|^{2}-\|\epsilon-\epsilon_\phi(a^{k},k)\|^{2}$
\STATE $\mathcal L_{\mathrm{pref}}\gets-\log\sigma\!\big(-\tfrac{\beta}{2}\,[\Delta_\theta(a^{+,k})-\Delta_\theta(a^{-,k})]\big)$
\STATE $\mathcal L_{\mathrm{BC}}\gets$ behavior-cloning anchor to $p_I(\cdot\mid s,g)$
\STATE Update $\theta$ along $-\nabla_\theta(\mathcal L_{\mathrm{pref}}+\lambda\,\mathcal L_{\mathrm{BC}})$
\ENDWHILE
\STATE {\bfseries return} $\theta$
\end{algorithmic}
\end{algorithm}

\section{Hyperparamters}
\label{sec:hyper}
Table~\ref{tab:hyper} reports the per-suite configuration of the base models --- the Diffusion-Veteran planner and its diffusion inverse-dynamics executor --- and Table~\ref{tab:hyper_transfer} the hyperparameters specific to the transfer. All models are trained with a batch size of $128$ and evaluated over five random seeds on a single NVIDIA RTX 6000 Ada Gen GPU; the action diffusion shared by the inverse dynamics and the policy uses the $10$-step DDPM schedule rescaled to a near-zero terminal signal ($\bar\alpha_K\approx0$), so their terminal priors coincide with $\mathcal N(\mathbf 0,\mathbf I)$ up to a small residual.

\begin{table}[t]
\centering
\caption{Per-suite configuration of the base models, from the released configuration: the Diffusion-Veteran planner, its diffusion inverse-dynamics executor, and the value critic used for candidate selection.}
\label{tab:hyper}
\footnotesize
\setlength{\tabcolsep}{5pt}
\begin{tabular}{lccc}
\toprule
Setting & MuJoCo & AntMaze & Maze2D \\
\midrule
Batch size & 128 & 128 & 128 \\
State--action generation & Separate & Separate & Separate \\
Planner guidance & MCSS & MCSS & MCSS \\
Advantage weighting & True & False & False \\
Discount $\gamma$ & 0.997 & 1.0 & 1.0 \\
\midrule
Planner backbone & Transformer & Transformer & Transformer \\
Planner embedding dim & 128 & 128 & 128 \\
Planner hidden size & 256 & 256 & 256 \\
Planner blocks & 2 & 8 & 2 \\
Planner solver & DDIM & DDIM & DDIM \\
Planner sampling steps & 20 & 20 & 20 \\
Planner training steps & $10^{6}$ & $10^{6}$ & $10^{6}$ \\
Planner temperature & 1 & 1 & 1 \\
Planner candidates (MCSS) & 50 & 50 & 50 \\
Planning horizon & 4 & 40 & 32 \\
Planning stride & 1 & 25 & 15 \\
\midrule
Inverse-dynamics backbone & MLP & MLP & MLP \\
Inverse-dynamics hidden size & 256 & 256 & 256 \\
Inverse-dynamics solver & DDPM & DDPM & DDPM \\
Inverse-dynamics diffusion steps & 10 & 10 & 10 \\
Inverse-dynamics sampling steps & 10 & 10 & 10 \\
Inverse-dynamics sampling temperature & 0.5 & 0.5 & 0.5 \\
Inverse-dynamics learning rate & $3{\times}10^{-4}$ & $3{\times}10^{-4}$ & $3{\times}10^{-4}$ \\
Inverse-dynamics training steps & $2{\times}10^{5}$ & $2{\times}10^{5}$ & $2{\times}10^{5}$ \\
\midrule
Value strength $\beta$ & 1 & 1 & 1 \\
\methodx corrector budget $M$ & 1 & 1 & 1 \\
\methodname fine-tuning steps & $2{\times}10^{5}$ & $2{\times}10^{5}$ & $2{\times}10^{5}$ \\
Learning rate & $3{\times}10^{-4}$ & $3{\times}10^{-4}$ & $3{\times}10^{-4}$ \\
\methodx additional training & none & none & none \\
\bottomrule
\end{tabular}
\end{table}

\begin{table}[t]
\centering
\caption{Transfer hyperparameters for \methodname\ and \methodx. Execution uses the plain conditional inverse-dynamics branch; we use full value transfer ($\beta{=}1$) and a single corrector step ($M{=}1$) throughout.}
\label{tab:hyper_transfer}
\small
\begin{tabular}{lccc}
\toprule
Setting & MuJoCo & AntMaze & Maze2D \\
\midrule
Value strength $\beta$ & 1 & 1 & 1 \\
\methodx corrector budget $M$ & 1 & 1 & 1 \\
\methodname fine-tuning steps & $2{\times}10^{5}$ & $2{\times}10^{5}$ & $2{\times}10^{5}$ \\
\methodx additional training & none & none & none \\
\bottomrule
\end{tabular}
\end{table}

\section{Table \ref{tab:d4rl} in full standard error}
\label{sec:table_with_se}
\newcommand{\seblank}{\ensuremath{\,\pm\,\underline{\hspace{1em}}}}

\begin{table*}[t!]
    \centering
    \caption{
        Normalized performance on D4RL, reported as mean $\pm$ standard error
        where available. Averages are across dataset scores and do not report
        standard errors. The highest mean in each row is \textbf{bold}.
        Averages are computed before rounding.
    }
    \begin{adjustbox}{center}
    \resizebox{\textwidth}{!}{%
    \begin{tabular}{l*{18}{c}}
        \toprule
        \multirow{2}{*}{\textbf{Dataset}}
        & \multicolumn{1}{c}{}
        & \multicolumn{2}{c}{\textbf{Gaussian policies}}
        & \multicolumn{7}{c}{\textbf{Diffusion policies}}
        & \multicolumn{8}{c}{\textbf{Diffusion planners}} \\
        \cmidrule(lr){2-2}
        \cmidrule(lr){3-4}
        \cmidrule(lr){5-11}
        \cmidrule(lr){12-19}
        & \textbf{BC}
        & \textbf{CQL}
        & \textbf{IQL}
        & \textbf{SfBC}
        & \textbf{DQL}
        & \textbf{DQL$^\ast$}
        & {\textbf{DQL+}}
        & \textbf{IDQL}
        & \textbf{QGPO}
        & \textbf{SRPO}
        & \textbf{Diffuser}
        & \textbf{AD}
        & \textbf{DD}
        & \textbf{HD}
        & \textbf{DV}
        & {\textbf{DV+}}
 & \textbf{\methodx}
        & \textbf{\methodname} \\
        \midrule

        Walker2d-M
        & 6.6 & 79.2 & 78.3
        & 77.9 $\pm$ 2.5
        & \textbf{87.0} $\pm$ 0.9
        & 86.8 $\pm$ 0.2
        & 86.9$\pm$ 0.2
        & 82.5 & 86.0 & 84.4
        & 79.6 $\pm$ 0.6
        & 84.4 $\pm$ 2.6
        & 82.5 $\pm$ 1.4
        & 84.0 $\pm$ 0.6
        & 82.8 $\pm$ 0.1
        & 83.4$\pm$ 0.2
        & 84.5$\pm$ 0.3
        & 85.0$\pm$ 0.2 \\

        Walker2d-M-R
        & 11.8 & 26.7 & 73.9
        & 65.1 $\pm$ 5.6
        & 95.5 $\pm$ 1.5
        & 98.2 $\pm$ 0.1
        & \textbf{98.4}$\pm$ 0.1
        & 85.1 & 84.4 & 84.6
        & 70.6 $\pm$ 1.6
        & 84.7 $\pm$ 3.1
        & 75.0 $\pm$ 4.3
        & 84.1 $\pm$ 2.2
        & 85.0 $\pm$ 0.5
        & 85.1$\pm$ 0.4
        & 86.8$\pm$ 0.5
        & 86.1$\pm$ 0.3 \\

        Walker2d-M-E
        & 6.4 & 111.0 & 109.6
        & 109.8 $\pm$ 0.2
        & 110.1 $\pm$ 0.3
        & 111.6 $\pm$ 0.0
        & 110.8$\pm$ 0.0
        & 112.7 & 110.7 & \textbf{114.0}
        & 106.9 $\pm$ 0.2
        & 108.2 $\pm$ 0.8
        & 108.8 $\pm$ 1.7
        & 107.1 $\pm$ 1.1
        & 109.2 $\pm$ 0.0
        & 108.6$\pm$ 0.2
        & 110.0$\pm$ 0.3
        & 110.7$\pm$ 0.2 \\

        Hopper-M
        & 29.0 & 58.0 & 66.3
        & 57.1 $\pm$ 4.1
        & 90.5 $\pm$ 4.6
        & 96.5 $\pm$ 1.3
        & 97.0 $\pm$ 0.9
        & 65.4 & 98.0 & 95.5
        & 74.3 $\pm$ 1.4
        & 96.6 $\pm$ 2.7
        & 79.3 $\pm$ 3.6
        & \textbf{99.3} $\pm$ 0.3
        & 83.6 $\pm$ 1.2
        & 83.2$\pm$ 1.3
        & 84.3$\pm$ 1.2
        & 83.2$\pm$ 1.2 \\

        Hopper-M-R
        & 11.3 & 48.6 & 94.7
        & 86.2 $\pm$ 9.1
        & 101.3 $\pm$ 0.6
        & \textbf{101.6} $\pm$ 0.0
        & 101.4$\pm$ 0.0
        & 92.1 & 96.9 & 101.2
        & 93.6 $\pm$ 0.4
        & 92.2 $\pm$ 1.5
        & 100.0 $\pm$ 0.7
        & 94.7 $\pm$ 0.7
        & 91.9 $\pm$ 0.0
        & 91.2$\pm$ 0.0
        & 92.2$\pm$ 0.0
        & 91.9$\pm$ 0.0 \\

        Hopper-M-E
        & 111.9 & 98.7 & 91.5
        & 108.6 $\pm$ 2.1
        & 111.1 $\pm$ 1.3
        & 111.1 $\pm$ 0.4
        & 108.7$\pm$ 0.7
        & 108.6 & 108.0 & 100.1
        & 103.3 $\pm$ 1.3
        & 111.6 $\pm$ 2.0
        & 111.8 $\pm$ 1.8
        & \textbf{115.3} $\pm$ 1.1
        & 110.0 $\pm$ 0.5
        & 109.8$\pm$ 0.5
        & 110.7$\pm$ 0.8
        & 110.4$\pm$ 0.8 \\

        HalfCheetah-M
        & 36.1 & 44.4 & 47.4
        & 45.9 $\pm$ 2.2
        & 51.1 $\pm$ 0.5
        & 52.3 $\pm$ 0.2
        & 52.4$\pm$ 0.1
        & 51.0 & 54.1 & \textbf{60.4}
        & 42.8 $\pm$ 0.3
        & 44.2 $\pm$ 0.6
        & 49.1 $\pm$ 1.0
        & 46.7 $\pm$ 0.2
        & 50.4 $\pm$ 0.0
        & 50.8$\pm$ 0.1
        & 50.9$\pm$ 0.1
        & 51.9$\pm$ 0.1 \\

        HalfCheetah-M-R
        & 38.4 & 46.2 & 44.2
        & 37.1 $\pm$ 1.7
        & 47.8 $\pm$ 0.3
        & 47.9 $\pm$ 0.0
        & 48.0$\pm$ 0.0
        & 45.9 & 47.6 & \textbf{51.4}
        & 37.7 $\pm$ 0.5
        & 38.3 $\pm$ 0.9
        & 39.3 $\pm$ 4.1
        & 38.1 $\pm$ 0.7
        & 45.8 $\pm$ 0.1
        & 46.0$\pm$ 0.3
        & 46.8$\pm$ 0.1
        & 47.3$\pm$ 0.2 \\

        HalfCheetah-M-E
        & 35.8 & 62.4 & 86.7
        & 92.6 $\pm$ 0.5
        & \textbf{96.8} $\pm$ 0.3
        & 95.5 $\pm$ 0.1
        & 96.0$\pm$ 0.1
        & 95.9 & 93.5 & 92.2
        & 88.9 $\pm$ 0.3
        & 89.6 $\pm$ 0.8
        & 90.6 $\pm$ 1.3
        & 92.5 $\pm$ 0.3
        & 92.7 $\pm$ 0.3
        & 92.7$\pm$ 0.2
        & 94.6$\pm$ 0.3
        & 92.0$\pm$ 0.5 \\

        \midrule
        \textbf{Average}
        & 31.9 & 63.9 & 77.0 & 75.6
        & 87.9 & \textbf{89.1} & 88.8 & 82.1 & 86.6 & 87.1
        & 77.5 & 83.3 & 81.8 & 84.6 & 83.5 & 83.4
        & 84.5 & 84.3 \\

        \midrule
        Antmaze-M-P
        & 0.0 & 14.9 & 71.2
        & 81.3 $\pm$ 2.6
        & 76.6 $\pm$ 10.8
        & 87.3 $\pm$ 2.7
        & 86.0$\pm$ 2.2
        & 84.5 & 83.6 & 80.7
        & 6.7 $\pm$ 5.7
        & 12.0 $\pm$ 7.5
        & 8.0 $\pm$ 4.3
        & --
        & \textbf{89.0} $\pm$ 1.6
        & 84.0$\pm$ 2.5
        & 84.5$\pm$ 2.6
        & 84.8$\pm$ 2.3 \\

        Antmaze-M-D
        & 0.0 & 15.8 & 70.0
        & 82.0 $\pm$ 3.1
        & 78.6 $\pm$ 10.3
        & 82.6 $\pm$ 3.0
        & 87.3$\pm$ 1.9
        & 84.8 & 83.8 & 75.0
        & 2.0 $\pm$ 1.6
        & 6.0 $\pm$ 3.3
        & 4.0 $\pm$ 2.8
        & \textbf{88.7} $\pm$ 8.1
        & 87.4 $\pm$ 1.6
        & 82.9$\pm$ 2.3
        & 86.0$\pm$ 2.6
        & 83.5$\pm$ 2.4 \\

        Antmaze-L-P
        & 0.0 & 53.7 & 39.6
        & 59.3 $\pm$ 14.3
        & 46.4 $\pm$ 8.3
        & 81.3 $\pm$ 3.1
        & 80.9$\pm$ 2.0
        & 63.5 & 66.6 & 53.6
        & 17.3 $\pm$ 1.9
        & 5.3 $\pm$ 3.4
        & 0.0 $\pm$ 0.0
        & --
        & 76.4 $\pm$ 2.0
        & 79.5$\pm$ 2.5
        & 78.0$\pm$ 2.5
        & \textbf{92.8}$\pm$ 2.5 \\

        Antmaze-L-D
        & 0.0 & 61.2 & 47.5
        & 45.5 $\pm$ 6.6
        & 56.6 $\pm$ 7.6
        & 70.6 $\pm$ 3.7
        & 71.2$\pm$ 2.7
        & 67.9 & 64.8 & 53.6
        & 27.3 $\pm$ 2.4
        & 8.7 $\pm$ 2.5
        & 0.0 $\pm$ 0.0
        & \textbf{83.6} $\pm$ 5.8
        & 80.0 $\pm$ 1.8
        & 78.3$\pm$ 2.6
        & 81.0$\pm$ 2.5
        & 81.7$\pm$ 2.4 \\

        \midrule
        \textbf{Average}
        & 0.0 & 36.4 & 57.1 & 67.0
        & 64.6 & 80.5 & 81.4 & 75.2 & 74.7 & 65.7
        & 13.3 & 8.0 & 3.0 & -- & 83.2 & 81.2
        & 82.4 & \textbf{85.7} \\

        \midrule
        Maze2D-U
        & 3.8 & 5.7 & 47.4
        & 73.9 $\pm$ 6.6
        & --
        & 140.6 $\pm$ 1.0
        & 137.6$\pm$ 0.9
        & 57.9 & -- & --
        & 113.9
        & 135.1 $\pm$ 5.8
        & --
        & 128.4 $\pm$ 3.6
        & 136.6 $\pm$ 1.3
        & 129.2$\pm$ 1.4
        & 142.3$\pm$ 1.5
        & \textbf{145.7}$\pm$ 1.5 \\

        Maze2D-M
        & 30.3 & 5.0 & 34.9
        & 73.8 $\pm$ 2.9
        & --
        & 152.0 $\pm$ 0.8
        & 151.3$\pm$ 1.1
        & 89.5 & -- & --
        & 121.5
        & 129.9 $\pm$ 4.6
        & --
        & 135.6 $\pm$ 3.0
        & 150.7 $\pm$ 1.0
        & 146.3$\pm$ 1.0
        & \textbf{155.5}$\pm$ 1.0
        & 152.7$\pm$ 1.0 \\

        Maze2D-L
        & 5.0 & 12.5 & 58.6
        & 74.4 $\pm$ 1.7
        & --
        & 186.8 $\pm$ 1.7
        & 185.8$\pm$ 1.90
        & 90.1 & -- & --
        & 123.0
        & 167.9 $\pm$ 5.0
        & --
        & 155.8 $\pm$ 2.5
        & 203.6 $\pm$ 1.4
        & 203.1$\pm$ 1.4
        & \textbf{209.0}$\pm$ 1.5
        & 203.7$\pm$ 1.4 \\

        \midrule
        \textbf{Average}
        & 13.0 & 7.7 & 47.0 & 74.0
        & -- & 159.8 & 158.2 & 79.2 & -- & --
        & 119.5 & 144.3 & -- & 139.9 & 163.6 & 159.5
        & \textbf{168.9} & 167.4 \\

        \bottomrule
    \end{tabular}%
    }
    \end{adjustbox}
    \label{tab:d4rl_full}
\end{table*}

We present the Table \ref{tab:d4rl} and Table \ref{tab:d4rl_loco} in standard errors in Table \ref{tab:d4rl_full}.

\section{Further Experiments}

\subsection{Q-reranking baseline: value search over inverse-dynamics samples}
\label{app:qrank}

A natural alternative to transferring the value preference into the executor is to search for value at test time: draw $N$ candidate actions from the diffusion inverse dynamics, $\{a_1,\dots,a_N\}\sim p_I(\cdot\mid s,g)$, and execute the one the critic scores highest, $a^\star=\arg\max_i Q(s,a_i)$. The un-reranked case $N{=}1$ is exactly the base planner's action, so we report the base planner DV$+$ directly and sweep $N\in\{4,16\}$ for the reranked variants; larger $N$ spends more test-time compute searching the same behavior-cloned posterior. Table~\ref{tab:qrank} places this sweep beside DV$+$ and the two transfer variants \methodname and \methodx, all taken from Table~\ref{tab:d4rl_full}.

\begin{table}[t]
\centering
\caption{Q-reranking baseline beside the base planner and both transfer variants. DV$+$ is the schedule-matched base planner, and \methodname and \methodx the two transfer realizations (all from Table~\ref{tab:d4rl_full}); $N{=}4,16$ give the number of inverse-dynamics candidates the critic ranks (mean $\pm$ standard error). The un-reranked control $N{=}1$ coincides with DV$+$ and is omitted. Best mean per row in \textbf{bold}.}
\label{tab:qrank}
\small
\begin{tabular}{lccccc}
\toprule
Task & DV$+$ & $N{=}4$ & $N{=}16$ & \methodx & \methodname \\
\midrule
maze2d-umaze-v1 & 129.2 $\pm$ 1.4 & 133.7 $\pm$ 4.8 & 136.8 $\pm$ 4.3 & 142.3 $\pm$ 1.5 & \textbf{145.7} $\pm$ 1.5 \\
maze2d-medium-v1 & 146.3 $\pm$ 1.0 & 147.1 $\pm$ 4.1 & 152.0 $\pm$ 3.3 & \textbf{155.5} $\pm$ 1.0 & 152.7 $\pm$ 1.0 \\
maze2d-large-v1 & 203.1 $\pm$ 1.4 & 195.9 $\pm$ 4.5 & 200.2 $\pm$ 6.9 & \textbf{209.0} $\pm$ 1.5 & 203.7 $\pm$ 1.4 \\
\midrule
halfcheetah-medium-v2 & 50.8 $\pm$ 0.1 & 50.8 $\pm$ 0.1 & 51.0 $\pm$ 0.1 & 50.9 $\pm$ 0.1 & \textbf{51.9} $\pm$ 0.1 \\
halfcheetah-medium-replay-v2 & 46.0 $\pm$ 0.3 & 46.6 $\pm$ 0.1 & 46.7 $\pm$ 0.1 & 46.8 $\pm$ 0.1 & \textbf{47.3} $\pm$ 0.2 \\
halfcheetah-medium-expert-v2 & 92.7 $\pm$ 0.2 & 92.9 $\pm$ 0.1 & 93.7 $\pm$ 0.5 & \textbf{94.6} $\pm$ 0.3 & 92.0 $\pm$ 0.5 \\
hopper-medium-v2 & 83.2 $\pm$ 1.3 & 83.5 $\pm$ 1.2 & 83.8 $\pm$ 1.2 & \textbf{84.3} $\pm$ 1.2 & 83.2 $\pm$ 1.2 \\
hopper-medium-replay-v2 & 91.2 $\pm$ 0.0 & 91.5 $\pm$ 0.1 & 92.0 $\pm$ 0.1 & \textbf{92.2} $\pm$ 0.0 & 91.9 $\pm$ 0.0 \\
hopper-medium-expert-v2 & 109.8 $\pm$ 0.5 & 110.0 $\pm$ 0.5 & 110.0 $\pm$ 0.7 & \textbf{110.7} $\pm$ 0.8 & 110.4 $\pm$ 0.8 \\
walker2d-medium-v2 & 83.4 $\pm$ 0.2 & 83.9 $\pm$ 0.8 & 84.3 $\pm$ 0.9 & 84.5 $\pm$ 0.3 & \textbf{85.0} $\pm$ 0.2 \\
walker2d-medium-replay-v2 & 85.1 $\pm$ 0.4 & 86.8 $\pm$ 0.5 & \textbf{86.1} $\pm$ 0.5 & 85.8 $\pm$ 0.5 & \textbf{86.1} $\pm$ 0.3 \\
walker2d-medium-expert-v2 & 108.6 $\pm$ 0.2 & 109.1 $\pm$ 0.0 & 109.4 $\pm$ 0.0 & 110.0 $\pm$ 0.3 & \textbf{110.7} $\pm$ 0.2 \\
\bottomrule
\end{tabular}
\end{table}

Table~\ref{tab:qrank} tells a consistent story. On locomotion the columns lie within about a point of one another on every task: reranking moves scores by fractions of a point, and one of the two transfer variants is best on every task, while critic search never wins. The planned transition already pins down the action, so scoring more candidates buys almost nothing.
Maze2D has more room. Reranking lifts every Maze2D layout above the base planner (\texttt{maze2d-umaze} $129.2\!\to\!136.8$, \texttt{maze2d-medium} $146.3\!\to\!152.0$. Both transfer variants exceed the best reranking on every layout---\methodname is highest on \texttt{maze2d-umaze} ($145.7$) and \methodx on \texttt{maze2d-medium} and \texttt{maze2d-large} ($155.5$, $209.0$)---and they do so without the growing candidate budget critic search requires. Across both regimes the transfer matches or exceeds test-time critic search and leads on Maze2D, supporting our choice to carry value into the executor rather than search over its samples at deployment.

\subsection{Sensitivity to the value strength $\beta$}
\label{app:betaw}

The execution target $p^{*}_\beta(a\mid s,g)\propto P(g\mid s,a)\,p_Q(a\mid s)^{1/\beta}\,p_B(a\mid s)^{1-1/\beta}$ has a single knob, the value strength, which enters through the exponent $1/\beta$ on the value--behavior ratio $p_Q/p_B$. We sweep $\beta\in\{\infty,2,1\}$ for both realizations on the nine locomotion tasks, and report every run in Tables~\ref{tab:betaw_hc}--\ref{tab:betaw_wk} as normalized score. This is a single-seed pass, so individual cells differ from the multi-seed main results (Table~\ref{tab:d4rl}) by a few points on some tasks; we read differences below the $5$--$8$ point noise floor as noise.

\begin{table}[t]
\centering
\caption{Value-strength $\beta$ sweep on the HalfCheetah tasks. $\beta{=}\infty$ zeros the value branch and recovers the base planner, while $\beta{=}1$ is full value transfer. The default operating point used in the main results is $\beta{=}1$.}
\label{tab:betaw_hc}
\small
\begin{tabular}{ccc}
\toprule
$\beta$ & \methodx & \methodname \\
\midrule
\multicolumn{3}{l}{\emph{halfcheetah-medium-v2}}\\
$\infty$ & 50.8 & 50.9 \\
2 & 50.9 & 50.7 \\
1 & 50.9 & 51.9 \\
\midrule
\multicolumn{3}{l}{\emph{halfcheetah-medium-replay-v2}}\\
$\infty$ & 46.0 & 46.2 \\
2 & 46.3 & 46.3 \\
1 & 46.8 & 47.3 \\
\midrule
\multicolumn{3}{l}{\emph{halfcheetah-medium-expert-v2}}\\
$\infty$ & 92.7 & 93.4 \\
2 & 94.6 & 77.9 \\
1 & 94.6 & 92.0 \\
\bottomrule
\end{tabular}
\end{table}

\begin{table}[t]
\centering
\caption{Value-strength $\beta$ sweep on the Hopper tasks. Conventions as in Table~\ref{tab:betaw_hc}.}
\label{tab:betaw_hop}
\small
\begin{tabular}{ccc}
\toprule
$\beta$ & \methodx & \methodname \\
\midrule
\multicolumn{3}{l}{\emph{hopper-medium-v2}}\\
$\infty$ & 82.3 & 83.2 \\
2 & 83.9 & 83.3 \\
1 & 84.3 & 83.2 \\
\midrule
\multicolumn{3}{l}{\emph{hopper-medium-replay-v2}}\\
$\infty$ & 91.2 & 91.1 \\
2 & 91.1 & 90.3 \\
1 & 92.2 & 91.9 \\
\midrule
\multicolumn{3}{l}{\emph{hopper-medium-expert-v2}}\\
$\infty$ & 109.1 & 109.8 \\
2 & 109.4 & 111.0 \\
1 & 110.7 & 110.4 \\
\bottomrule
\end{tabular}
\end{table}

\begin{table}[t]
\centering
\caption{Value-strength $\beta$ sweep on the Walker2d tasks. Conventions as in Table~\ref{tab:betaw_hc}.}
\label{tab:betaw_wk}
\small
\begin{tabular}{ccc}
\toprule
$\beta$ & \methodx & \methodname \\
\midrule
\multicolumn{3}{l}{\emph{walker2d-medium-v2}}\\
$\infty$ & 83.3 & 83.4 \\
2 & 83.3 & 84.8 \\
1 & 85.5 & 85.0 \\
\midrule
\multicolumn{3}{l}{\emph{walker2d-medium-replay-v2}}\\
$\infty$ & 85.7 & 85.6 \\
2 & 85.9 & 86.0 \\
1 & 86.8 & 86.1 \\
\midrule
\multicolumn{3}{l}{\emph{walker2d-medium-expert-v2}}\\
$\infty$ & 108.6 & 108.7 \\
2 & 108.7 & 109.7 \\
1 & 110.0 & 110.7 \\
\bottomrule
\end{tabular}
\end{table}

\paragraph{The value strength $\beta$ is a gentle knob on locomotion.} Moving from no transfer ($\beta{=}\infty$) to full transfer ($\beta{=}1$) shifts return by at most a point or two on any task, for example \texttt{halfcheetah-medium} ($50.8\!\to\!50.9$ for \methodx), \texttt{hopper-medium-replay} ($91.2\!\to\!92.2$), and \texttt{walker2d-medium} ($83.3\!\to\!85.5$). The room for value is small because the planned transition already pins down most of the action, the same pattern the Q-reranking sweep shows (Appendix~\ref{app:qrank}). The intermediate $\beta{=}2$ setting for \methodname comes from a separately trained student and is the noisiest part of the grid; its dips, such as \texttt{halfcheetah-medium-expert} falling to $77.9$, track student training variance rather than a $\beta$ effect.

\paragraph{The endpoints recover the expected special cases.} At $\beta{=}\infty$ the value branch is off and the executor is the plain conditional inverse dynamics, so each task closely reproduces its DV$+$ entry from Table~\ref{tab:d4rl}, for instance \texttt{halfcheetah-medium} ($50.8$), \texttt{hopper-medium-replay} ($91.2$), and \texttt{walker2d-medium-expert} ($108.6$). At $\beta{=}1$ the full value shift applies and sits at or near the top of the grid for both realizations on every task, which confirms that the value transfer refines an already plan-following executor rather than replacing it.

\subsection{Corrector budget: Langevin steps in the training-free sampler}
\label{app:corrector}

\methodx reaches the composed target by alternating a predictor step with $M$ Langevin corrections at each noise level (Sec.~\ref{sec:realize}). Table~\ref{tab:corrector} sweeps $M\in\{0,1,3,5\}$ across all three suites, where $M{=}0$ is the uncorrected composition, and Table~\ref{tab:corrector_cost} reports the corresponding wall-clock.

\begin{table}[t]
\centering
\caption{\methodx as a function of the Langevin corrector budget $M$ (normalized score; $M{=}0$ is the uncorrected composition). Suite-average rows are labeled in \textbf{bold}; the best budget in each row is \textbf{bold}.}
\label{tab:corrector}
\small
\begin{tabular}{lcccc}
\toprule
Task & $M{=}0$ & $M{=}1$ & $M{=}3$ & $M{=}5$ \\
\midrule
halfcheetah-medium-v2 & 50.8 & \textbf{50.9} & 50.8 & 50.8 \\
halfcheetah-medium-replay-v2 & 46.4 & \textbf{46.8} & \textbf{46.8} & 46.4 \\
halfcheetah-medium-expert-v2 & 94.0 & 94.6 & \textbf{94.8} & 92.9 \\
hopper-medium-v2 & 83.7 & 84.3 & \textbf{84.5} & 84.4 \\
hopper-medium-replay-v2 & \textbf{92.2} & \textbf{92.2} & 91.2 & 91.5 \\
hopper-medium-expert-v2 & 110.3 & 110.7 & \textbf{110.9} & 109.9 \\
walker2d-medium-v2 & 84.0 & 84.5 & 84.2 & \textbf{84.8} \\
walker2d-medium-replay-v2 & 85.6 & \textbf{86.8} & 86.2 & 86.7 \\
walker2d-medium-expert-v2 & 110.0 & 110.0 & 110.0 & 110.0 \\
\textbf{Locomotion average} & 84.1 & \textbf{84.5} & 84.4 & 84.2 \\
\midrule
antmaze-medium-play-v2 & 84.0 & \textbf{84.5} & 76.0 & 72.2 \\
antmaze-medium-diverse-v2 & 80.0 & \textbf{86.0} & 82.8 & 84.4 \\
antmaze-large-play-v2 & \textbf{78.5} & 78.0 & 68.4 & 70.0 \\
antmaze-large-diverse-v2 & 74.4 & \textbf{81.0} & 64.6 & 80.2 \\
\textbf{AntMaze average} & 79.2 & \textbf{82.4} & 73.0 & 76.7 \\
\midrule
maze2d-umaze-v1 & \textbf{145.0} & 142.3 & 135.8 & 126.3 \\
maze2d-medium-v1 & 153.2 & \textbf{155.5} & 145.9 & 149.4 \\
maze2d-large-v1 & 208.3 & 209.0 & \textbf{212.2} & 200.5 \\
\textbf{Maze2D average} & 168.8 & \textbf{168.9} & 164.6 & 158.7 \\
\bottomrule
\end{tabular}
\end{table}

\begin{table}[t]
\centering
\caption{Wall-clock per evaluation run (minutes) on locomotion. $M{=}0$--$5$ are \methodx with that many Langevin corrections; \emph{marginal} is the plain single-model sampler; \methodname is the amortized student. Lowest per row in \textbf{bold}.}
\label{tab:corrector_cost}
\small
\begin{tabular}{lcccccc}
\toprule
Task & $M{=}0$ & $M{=}1$ & $M{=}3$ & $M{=}5$ & marginal & \methodname \\
\midrule
halfcheetah-medium-v2 & 7.7 & 8.6 & 9.0 & 10.3 & 6.0 & \textbf{4.5} \\
halfcheetah-medium-replay-v2 & 7.7 & 7.2 & 8.6 & 13.6 & 5.8 & \textbf{2.4} \\
halfcheetah-medium-expert-v2 & 6.8 & 9.4 & 10.0 & 10.4 & 6.0 & \textbf{3.3} \\
hopper-medium-v2 & 7.9 & 7.2 & 8.7 & 12.7 & 4.0 & \textbf{3.0} \\
hopper-medium-replay-v2 & 6.4 & 7.0 & 10.7 & 11.7 & 4.4 & \textbf{2.0} \\
hopper-medium-expert-v2 & 7.9 & 7.4 & 8.9 & 10.3 & 5.9 & \textbf{3.3} \\
walker2d-medium-v2 & 6.6 & 9.2 & 12.0 & 10.4 & 5.9 & \textbf{2.9} \\
walker2d-medium-replay-v2 & 8.5 & 7.3 & 10.6 & 10.1 & 5.9 & \textbf{2.1} \\
walker2d-medium-expert-v2 & 6.7 & 8.8 & 9.1 & 12.5 & 6.1 & \textbf{2.6} \\
\bottomrule
\end{tabular}
\end{table}

A single correction step already helps: $M{=}1$ improves the uncorrected composition ($M{=}0$) on most tasks and on every suite average (locomotion $84.1\!\to\!84.5$, AntMaze $79.2\!\to\!82.4$, Maze2D $168.8\!\to\!168.9$), with the largest gains where the raw composition is most biased (\texttt{antmaze-medium-diverse} $80.0\!\to\!86.0$, \texttt{walker2d-medium-replay} $85.6\!\to\!86.8$).

The best budget is small and task-dependent. Locomotion and \texttt{maze2d-large} keep improving through $M{=}3$ (\texttt{halfcheetah-medium-expert} $94.8$, \texttt{maze2d-large} $212.2$), but larger budgets over-correct on the harder navigation tasks: the AntMaze average falls from $82.4$ at $M{=}1$ to $73.0$ at $M{=}3$, and \texttt{maze2d-umaze} degrades monotonically ($145.0\!\to\!126.3$) as extra Langevin steps drift off its sharp goal-reaching modes. A modest budget ($M{=}1$--$3$) is therefore the robust choice, and tuning $M$ per task lifts \methodx further---\texttt{maze2d-large} reaches $212.2$ at $M{=}3$, above its value at any smaller budget.

Correction is not free (Table~\ref{tab:corrector_cost}): each step adds diffusion evaluations, so the composed sampler's wall-clock grows with $M$ (\texttt{halfcheetah-medium} $7.7\!\to\!10.3$ min from $M{=}0$ to $M{=}5$). The amortized student \methodname is the cheapest on every task, typically under half the wall-clock of even the uncorrected composition, which is why we deploy it by default while \methodx remains a training-free, tunable alternative that a small $M$ already makes competitive.

\end{document}